\documentclass{article}

\usepackage[final,nonatbib]{neurips_data_2024}

\usepackage[utf8]{inputenc} % allow utf-8 input
\usepackage[T1]{fontenc}    % use 8-bit T1 fonts
\usepackage{hyperref}       % hyperlinks
\usepackage{url}            % simple URL typesetting
\usepackage{booktabs}       % professional-quality tables
\usepackage{amsfonts}       % blackboard math symbols
\usepackage{nicefrac}       % compact symbols for 1/2, etc.
\usepackage{microtype}      % microtypography
\usepackage{xcolor}         % colors

\usepackage{lipsum}		    
\usepackage{graphicx}
\usepackage{doi}
\usepackage{caption} 
\usepackage{xspace}
\usepackage{subcaption}
\usepackage{array}
\usepackage{mdframed}
\usepackage{listings}
\usepackage{multirow}
\usepackage{xcolor}
\usepackage{booktabs}
\usepackage{tabularx}
\usepackage{amssymb}
\usepackage{pifont}
\usepackage[affil-it]{authblk}
\usepackage{wrapfig}

\usepackage{amsmath}

\newcommand{\prompt}{ER\xspace}
\newcommand{\dataset}{\textsc{AndroidControl}\xspace}

\newcommand{\tasks}{\textsc{15,283}\xspace}
\newcommand{\insts}{\textsc{14,548}\xspace}
\newcommand{\apps}{\textsc{833}\xspace}
\newcommand{\steps}{\textsc{5.5}\xspace}
\newcommand{\id}[1]{{\sf\small #1}}

\newcommand\down[1]{{\color{red}{[#1]}}}
\newcommand\up[1]{{\color{green}{[#1]}}}

\newcommand{\cmark}{\textcolor{green}{\ding{51}}} % Green check mark
\newcommand{\xmark}{\textcolor{red}{\ding{55}}} % Red cross

\iftrue
\newcommand{\exclude}[1]{}
\else
\newcommand{\exclude}[1]{#1}
\fi

\title{On the Effects of Data Scale on \\ UI Control Agents}
\author[1]{Wei Li}
\author[1]{William Bishop}
\author[1]{Alice Li}
\author[1]{Chris Rawles}
\author[1]{Folawiyo Campbell-Ajala}
\author[2]{\\Divya Tyamagundlu}
\author[1]{Oriana Riva}
\affil[1]{Google DeepMind}
\affil[2]{Google}

\begin{document}
\maketitle

\begin{abstract}
  Autonomous agents that control user interfaces to accomplish human tasks are emerging. Leveraging LLMs to power such agents has been of special interest, but unless fine-tuned on human-collected task demonstrations, performance is still relatively low. In this work we study whether fine-tuning alone is a viable approach for building real-world UI control agents. %In particularly, we investigate how performance measured on both high and low-level tasks in domain and out of domain scales as more training data is collected.
  To this end we collect and release a new dataset, \dataset, consisting of \tasks demonstrations of everyday tasks with Android apps. Compared to existing datasets, each \dataset task instance includes both high and low-level human-generated instructions, allowing us to explore the level of task complexity an agent can handle. Moreover, \dataset is the most diverse UI control dataset to date, including \insts unique tasks over \apps Android apps, thus allowing us to conduct in-depth analysis of the model performance in and out of the domain of the training data. Using the dataset, we find that when tested in domain fine-tuned models outperform zero and few-shot baselines and scale in such a way that robust performance might feasibly be obtained simply by collecting more data.  Out of domain, performance scales significantly more slowly and suggests that in particular for high-level tasks, fine-tuning on more data alone may be insufficient for achieving robust out-of-domain performance. 
\end{abstract}

\section{Introduction}\label{sec:introduction}

Recent work has studied how large language models (LLMs) can be leveraged to build UI control agents~\cite{koh2024visualwebarena,zheng2023seeact,yan2023gpt4v,rci_kim2023language,deng2023mind2web} that accomplish human tasks by interacting with a digital device environment. These agents perceive the state of the device by observing its screen (from screenshots or application UI trees), and generate actions (click, type, scroll, etc.) that are executed through the device's user interface. Tasks, specified in natural language, can range from configuring device settings and sending emails, to navigating shopping websites and planning a trip. 

While progress is rapidly advancing, absolute performance of UI control agents that leverage pre-trained LLMs without fine-tuning on task demonstrations is still relatively low.  When tested in real-world environments, where agents control everyday applications and websites, recently-reported task success rates range from 12\% on desktop applications~\cite{xie2024osworld} to 46\% on mobile applications~\cite{bishop2024latent}. In contrast, agents that leverage models fine-tuned for task execution~\cite{nakano21:webgpt,webagent:iclr2024}, achieve even 80\%~\cite{webagent:iclr2024} success rate, when tested on websites and tasks similar to what they are trained on.  

While the pattern of collecting new datasets and fine-tuning shows promise, there are at least two important unanswered questions. First, to the best of our knowledge no prior work has examined the question of scaling: how much data must be collected in order to obtain a given performance level with fine-tuned models. This question is particularly important because human demonstrations of UI interactions for fine-tuning are time consuming and expensive to collect. Understanding how performance scales, both in domain and out of the domain of the collected demonstrations (unseen tasks and unseen applications), is important for determining whether fine-tuning alone is a viable path towards deploying UI control agents in the real world. Therefore, one of the main goals of this work is to rigorously quantify how performance of fine-tuned agents scales, both in and out of domain, as the amount of data used for fine-tuning is increased. 

Second, it is not clear the level of task complexity fine-tuning might be fruitfully applied to. Conceptually, UI control agents must both decompose a high-level goal into a set of small atomic actions and execute (``ground'') those actions in a device screen.  While the high-level reasoning with LLMs, required for determining how to accomplish high-level goals, is still an open problem in artificial intelligence~\cite{wei2021finetuned,wang2022self,yao2022,zhou2022least,kojima-nips22,yao2023, wei2023chainofthought}, the set of low-level actions (clicking, typing, etc...) required to execute tasks are more constrained, and general agents capable of robust grounding across domains might be approachable via fine-tuning.  Therefore, a second goal of this work is to quantify the scaling of fine-tuning for agents performing both high-level and low-level tasks. 

\begin{figure}[t]
\centering
\includegraphics[width=0.97\textwidth]{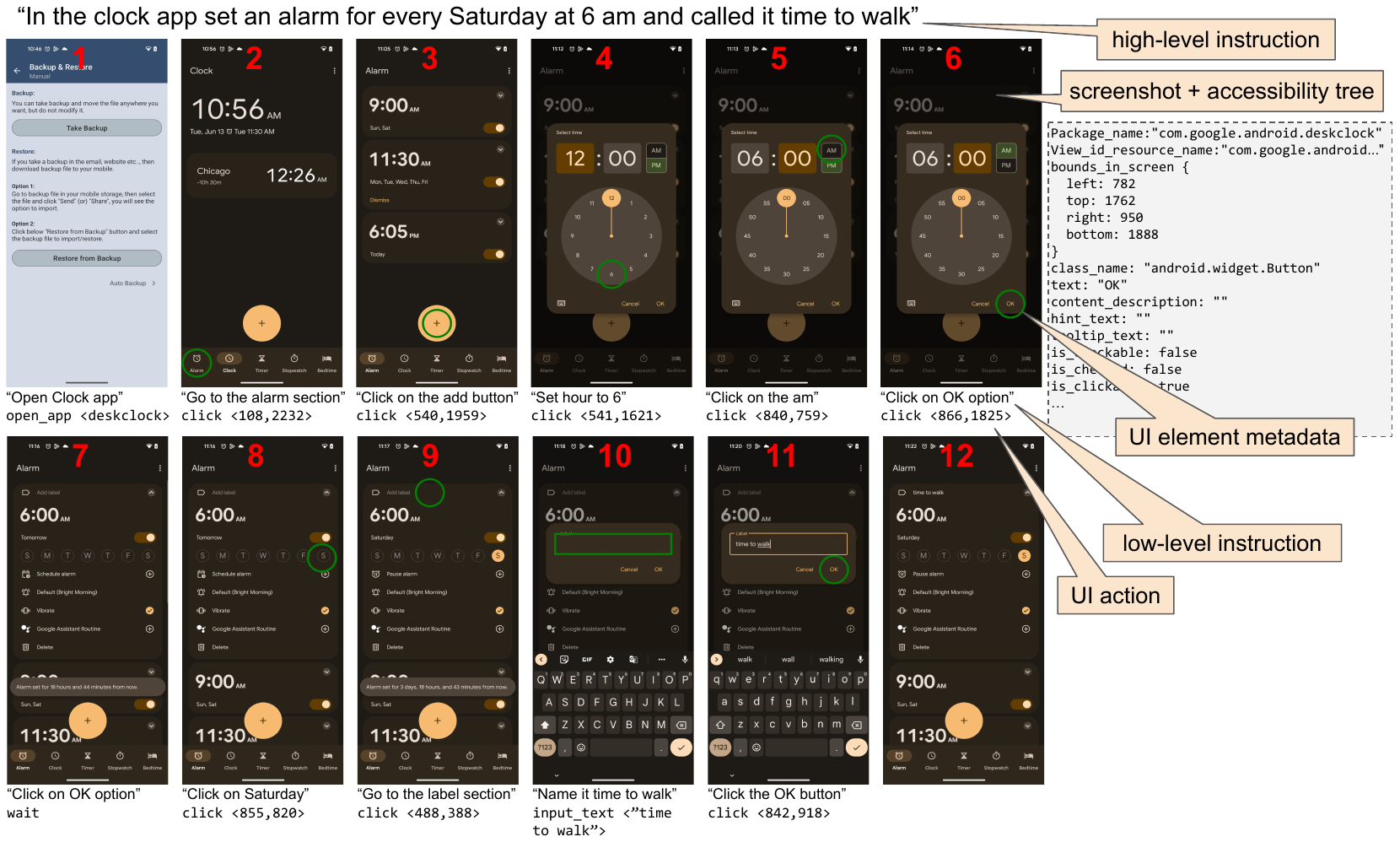}
\vspace{-0.7ex}
\caption{An example task demonstration contained in \dataset. Green circles/rectangles highlight the action on the screen. Red numbers are added only for illustration purposes.}
\label{fig:overview}
\end{figure}

Rigorously quantifying scaling in these ways requires a carefully constructed dataset.  To this end, we introduce \dataset, a large-scale dataset of \tasks demonstrations of tasks performed by humans in Android apps. Figure~\ref{fig:overview} shows an example data sample. Compared to existing datasets~\cite{deng2023mind2web,aitw2023}, for every task \dataset provides both the high- and low-level human-generated instructions describing it. This is essential to investigate the level of task complexity a model can handle and also provides richer supervision during training. \dataset is also the most diverse UI control dataset that exists today, including \insts unique tasks over \apps different Android apps, thus allowing us to generate multiple test splits for measuring performance in and out of domain. % This combination of characteristics makes it an ideal dataset for assessing the scaling questions we seek to answer. 
As a resource to the community, we make \dataset publicly available.\footnote{\url{https://github.com/google-research/google-research/tree/master/android_control}}

Overall, we make the following contributions: \emph{(i)} we collect and release \dataset, a new UI control dataset whose size, structure and diversity advances previous datasets, \emph{(ii)} we use \dataset to quantify how fine-tuning with demonstrations scales when applied to both low- and high-level tasks and to tasks in and out of the domain of the training data, and \emph{(iii)} we compare fine-tuning to various zero-shot and few-shot baselines, finding that fine-tuning scales favorably in domain, but out of domain, it requires one or two orders of magnitude more data to obtain robust performance on high-level tasks, suggesting that additional approaches may be beneficial for obtaining agents which robustly perform out-of-domain high-level tasks.

\section{Related work}

\paragraph{UI control datasets}

\begin{table*}[t]
	\centering
	\scalebox{0.71}{
	\begin{tabular}{llrrrccccc}
		\toprule
		Dataset &  Platform & \multicolumn{1}{c}{\# Human} & \multicolumn{1}{c}{\# Unique} & \# Apps or  & \# Task & \multicolumn{1}{c}{UI tree?} & \multicolumn{1}{c}{Screen?} & High-level & Low-level\\
		& & \multicolumn{1}{c}{demos} & \multicolumn{1}{c}{instr.} & websites & steps & & & instr. & instr. \\
		\midrule
		MiniWoB++~\cite{miniwob} & Web (synthetic) & 17,971 & 100 & 114  & 2.3 & \cmark & \xmark & \xmark & \cmark \\
		WebShop~\cite{yao2023webshop} & Web &  1,566  & 1,566  &  1 & 11.3~~  & \xmark & \cmark  &   \cmark  &  \xmark     \\
        UIBert~\cite{bai2021uibert} & Android  & 16,660 & - & -   & 1.0 & \cmark & \cmark  & \xmark &   \cmark \\
        PixelHelp~\cite{li-acl20}  & Android  & 187 & 187 &4   & 4.2 & \cmark & \xmark &   \cmark  &   \cmark\\
        UGIF~\cite{ugif}  &  Android  & 523 & 523  & 12  & 5.3 & \cmark & \cmark &  \cmark &   \cmark \\
        MoTIF~\cite{motif}  & Android & 4,707 & 276 &  125   & 4.5 & \cmark & \cmark &   \cmark &   \cmark  \\
        Mind2Web~\cite{deng2023mind2web}  & Web  & 2,350  & 2,350 &  137   & 7.3 & \cmark & \cmark &   \cmark & \xmark  \\
        AitW~\cite{aitw2023} & Android &  715,142 & 30,378 &  357  & 6.5 & \xmark & \cmark & \cmark & 
        \xmark       \\
        WebVoyager~\cite{he2024webvoyager} & Web & 643  & 643 &  15  & - & \cmark & \cmark &   \cmark  & \xmark    \\
        WebLINX~\cite{lu2024weblinx} & Web &  2,337  & 2,377 & 155  & 43.0 & \cmark & \cmark &  \cmark & \xmark \\
        \midrule
        \textbf{\dataset}  & Android & \tasks  & \insts & \apps  & \steps & \cmark & \cmark &   \cmark &  \cmark     \\
		\bottomrule
	\end{tabular}
	}
    \caption{Comparison of \dataset to existing UI control datasets. We consider device platform, size (as number of task demonstrations), diversity (as number of unique task instructions and apps/websites), average number of steps in a task, how the UI state is captured (UI tree vs. screenshot), and whether tasks are described through high-level instructions or sequences of low-level commands.}
    \label{table:dataset-comparison}
\end{table*}

Table~\ref{table:dataset-comparison} compares \dataset to existing UI control datasets. The structure of these datasets is similar. They consist of a natural language task description and a human-recorded demonstration, in the form of a sequence of UI actions (click, type, swipe, etc.) and associated UI states. 
%A UI action contains information about the associated UI element or any other input argument. %UI states are represented by screenshots or UI trees. 
What differentiates these datasets is mainly whether they are single-step (as in grounding referring expressions datasets like UIBert~\cite{bai2021uibert}), whether the task description is expressed as a high-level goal or as a sequence of low-level step instructions, and how the UI state is represented (screenshot vs. UI tree). Three features make \dataset unique. First, for every task, it contains both low-level and high-level instructions generated by human annotators. While a few other datasets contain both these types of annotation, their low-level instructions are either synthetically generated (as in MoTIF~\cite{motif}) or are limited to one action type (only click actions~\cite{li-acl20,ugif}). In addition to bringing richer language supervision during training, the availability of human-generated low-level instructions allows us to test UI control agents on different levels of task complexity. Second, if we consider the number of unique task instructions and the number of human demonstrations, \dataset is the second-largest UI control dataset to date, second only to AitW~\cite{aitw2023}.\footnote{AitW is heavily web focused. If we consider the AitW episodes involving Android apps, we obtain $~$505k episodes across 167 apps, for a total of 1.6k unique task instructions, a much smaller number of \dataset. AitW also does not contain application UI trees, thus making the UI state representations incomplete.} The diversity of task scenarios present in \dataset is its third differentiating feature: \dataset includes tasks from \apps different Android apps, 6 times more than popular datasets like Mind2Web~\cite{deng2023mind2web} and almost 5 times more than AitW. This diversity makes \dataset optimal for realistic, out-of-domain analysis. Note that Mind2Web also provides out-of-domain splits but given its smaller size (2,350 tasks over 137 websites, with a train split of 1k demonstrations) is not suitable for a scaling analysis.

In addition to the datasets listed in Table~\ref{table:dataset-comparison}, recent work proposes interactive testing environments for UI control agents~\cite{miniwob, zhou2024webarena, koh2024visualwebarena, xie2024osworld, android_world,bonatti2024windowsagentarenaevaluating} where the environment provides the agents with reward signals. These environments are designed for online testing and are limited to no more than 20 applications or websites. The only exception is MiniWob~\cite{miniwob} for which task demonstrations have been collected, but the environment consists of much simplified, synthetic websites.

\paragraph{UI control agents}
Early UI control agents were trained from scratch using behavioural cloning~\cite{pmlr-v162-humphreys22a,uinav-sys,li-acl20} or reinforcement learning~\cite{web-workflows18,learning-navigate-web}. Current UI agents use pre-trained LLMs and multimodal models. One line of work prompts LLMs in a zero-shot or few-shot regime~\cite{yan2023gpt4v,aitw2023,he2024webvoyager,koh2024visualwebarena,rci_kim2023language, zheng2023synapse}. Another line of work relies on fine-tuning which is applied end to end~\cite{nakano21:webgpt} or to build specific model capabilities, such as identifying the interactable UI elements in a webpage~\cite{zheng2023seeact,webagent:iclr2024,deng2023mind2web}.
To name a few, SeeAct~\cite{zheng2023seeact}, which we use in our evaluation, is a web agent that leverages large multimodal models to understand text and visual elements on webpages. The best-performing SeeAct agent relies on a fine-tuned cross-encoder model to select candidates web elements for interaction. WebGPT~\cite{nakano21:webgpt} fine-tunes GPT-3 to learn to use a web browser. WebAgent~\cite{webagent:iclr2024} pre-trains a T5 model to extract HTML snippets and leverages Flan-U-PaLM to generate Python code to control a web environment. Synapse~\cite{zheng2023synapse} introduces a trajectory-as-exemplar prompting method where memory of previous interactions allows the agent to perform complex, multi-step tasks. 

\paragraph{Domain generalization}
As evidenced by various LLM studies~\cite{gpt3,kaplan2020scaling,hernandez2021scaling}, scaling model and data size for the training leads to steady improvements in domain generalization. On the other hand, when transferring a pre-trained model to a downstream task through fine-tuning, while in-distribution performance improves, a reduction in the robustness to distribution shifts is observed~\cite{kumar2022finetuning,pmlr-v162-wortsman22a, andreassen2021evolution}. In this work, we empirically study how scaling data size in fine-tuning affects in-domain and out-of-domain performance of UI control agents. While prior work has tested UI agents using out-of-domain test splits~\cite{deng2023mind2web,motif}, to the best of our knowledge a data scale analysis has not been conducted. To minimize training cost and to maintain the out-of-domain generalization, we evaluate also the option of not fine-tuning, by evaluating multiple zero-shot and few-shot baselines.

\section{The \dataset dataset}
\label{sec:dataset}

The collection of the \dataset dataset is motivated by our dual goal of studying (i) how scaling data size for fine-tuning UI control models affects in-domain and out-of-domain performance, and (ii) the level of task complexity these fine-tuned models can be effective for. 

\subsection{Data collection}
\label{sec:data_collection}

\begin{wrapfigure}{r}{0.45\textwidth}
\includegraphics[width=\linewidth]{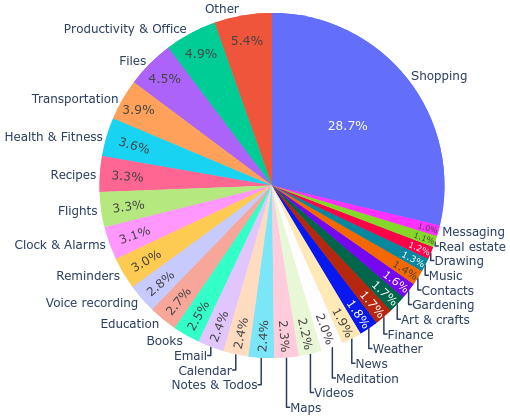}
\caption{Distribution of the app categories that compose \dataset.}
\label{fig:categories}
\vspace{-2ex}
\end{wrapfigure}

We collect \dataset using crowdsourcing over the course of a year. The data collection starts by giving crowdworkers generic feature descriptions for apps from 40 different categories (Figure~\ref{fig:categories}). These descriptions are generated using LLMs (e.g., "in a note taking app you can create a new note with details"). Then, we ask crowdworkers to instantiate each feature description into one or multiple tasks involving apps of their choice.

By allowing annotators to use any app of their choice we succeed in collecting a largely-varied dataset encompassing \apps Android apps, including Google apps (Settings, Gmail, Google Maps, etc.), high-trend apps (e.g., Amazon, Booking.com, Kayak, Spotify, etc.) as well as less-popular or regional apps. This is important because high-popularity apps tend to include well-annotated accessibility trees and have more user-friendly interfaces, thus possibly facilitating the agent's task. We confirm this assumption by analyzing the performance of some of our tested agents on Google apps and non-Google apps (see results in Section~\ref{app:app-types} in the Appendix).

During collection of a demonstration, annotators first provide a high-level description of a task in natural language (e.g., "Add an alarm to wake me up on Saturday mornings at 6am"). We ask annotators to make the descriptions detailed enough to be interpretable without any ambiguity. We also instruct them to always include the name of the target app in the task description, unless obvious (e.g., Google first-party apps such as Clock or Settings). By doing so, the collected data can enable us to test memory-less, single-turn agent interactions. 

\begin{figure}
  \begin{minipage}[b]{.57\linewidth}
    \centering
    \scalebox{0.74}{
    \begin{tabular}{ll}
    \toprule
    \id{click <elem>}  & Click the center of the specified element.\\
    \id{long\_press <elem>} & Long press the center of the specified element.\\
    \id{input\_text <text>} & Type the specified text. \\
    \id{scroll} <direction> & Scroll in the specified direction.\\
    \midrule
    \id{navigate\_home} & Go to the home screen.\\
    \id{navigate\_back} & Go back to the previous app screen. \\
    \id{open\_app <app>} & Launch the specified app.\\
    \id{wait} & Wait until the next observation is received.\\
    \bottomrule
    \end{tabular}
    }
    \vspace{-1ex}
    \captionof{table}{Actions captured in \dataset.}
    \label{tab:action-space}
  \end{minipage}\hfill
  \begin{minipage}[b]{.37\linewidth}
    \centering
     \includegraphics[width=\linewidth]{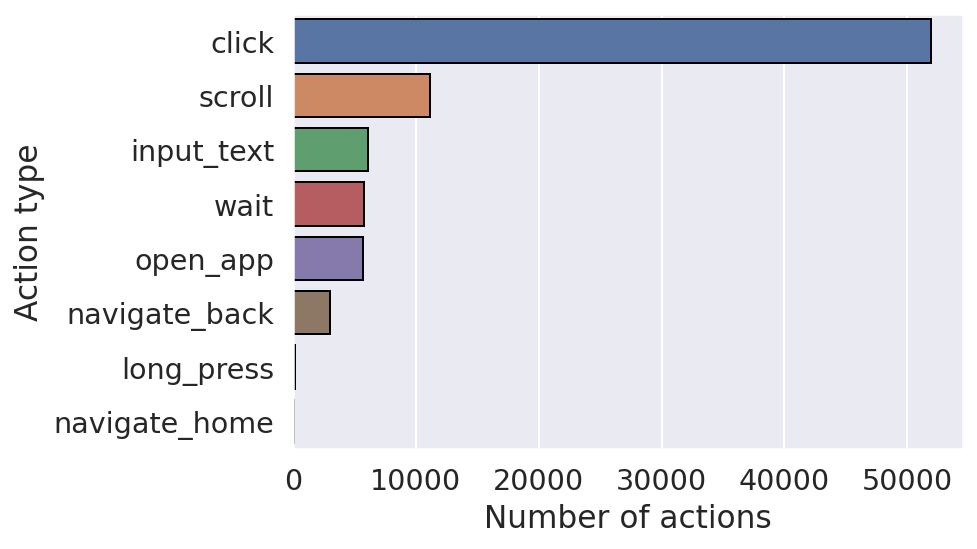}
     \vspace{-4ex}
     \caption{Action distribution}
     \label{fig:elements}
  \end{minipage}
\end{figure}

In order to collect interaction traces, each annotator is provided with a setup that includes a physical Android phone (Google Pixel with Android 8.0 or higher) installed with a companion Android app that in turn connects to a web app running on a desktop Chrome browser. Annotators control the phone through the web app, using the WebUSB protocol and Android Debug Bridge (ADB). The web app provides annotators with controls to perform actions on the phone and observe their outcome. An annotator can select from the following set of UI actions to perform on the phone: \id{click}, \id{long\_press}, \id{input\_text}, \id{scroll}, \id{navigate\_home}, \id{navigate\_back}, \id{open\_app} and \id{wait} (see Table~\ref{tab:action-space}). For each action, applicable metadata such as touch coordinates, target elements, entered text, and timing information are automatically appended to the interaction trace (see Appendix~\ref{app:dataset} for more details). Annotators are instructed to avoid performing actions that are unnecessary or unrelated to the task. After an action is executed, a real-time screenshot of the phone’s display is shown to the annotator and added to the interaction trace. This enables the annotator to completely operate their phone through the web app.  Before executing each  action, the annotator is asked to type in a short natural language description of the action they are about to take ("add a new alarm", "set the hours to 6", etc.), as if they were instructing someone to execute that action.  These are also incorporated into the interaction trace and make up the low-level instructions in the dataset. If annotators realize the task is not feasible because of an unsupported functionality in the app or because of an error they tag the trace as \id{infeasible} or \id{failed}, respectively. Otherwise, it is tagged as \id{successful}.

As both high-level and low-level instructions are generated by annotators, they may contain typo and grammatically errors (for example, the high-level instruction in Figure~\ref{fig:overview}). We left such errors in the dataset because humans often produce errors of this type.

Overall, this data collection involved 20 annotators. Each annotator went through a training process of several weeks. To maximize the diversity of the task demonstrations, in the last 4 months of the data collection we asked annotators to impersonate 20 different personas. Persona profiles are generated using an LLM prompted with detailed ranging from name, address, occupation, hobbies to upcoming week-end plans, family relationships, and typical day schedule.

%\begin{table}[t]
%\centering
%\caption{Actions captured in \dataset: \id{click}, \id{long\_press} and \id{input\_text} take a target element as input; \id{scroll} requires a direction. Four special actions, \id{navigate\_home}, \id{navigate\_back}, \id{open\_app} and \id{wait}, are executed using ADB.}
%\scalebox{0.8}{
%\begin{tabular}{ll}
%\toprule
%\id{click <elem>}  & Click the center of the specified element.\\
%\id{long\_press <elem>} & Long press the center of the specified element.\\
%\id{input\_text <text,elem>} & Type the text in the specified element. \\
%\id{scroll} <direction> & Scroll in the specified direction.\\
%\midrule
%\id{navigate\_home} & Go to the home screen.\\
%\id{navigate\_back} & Go back to the previous app screen. \\
%\id{open\_app <app\_name>} & Launch the specified application.\\
%\id{wait} & Wait until the next observation is received.\\
%\bottomrule
%\end{tabular}
%}
%\end{table}

\begin{figure*}[t]
\begin{subfigure}[t]{0.33\textwidth}
    \centering
    \includegraphics[width=\textwidth]{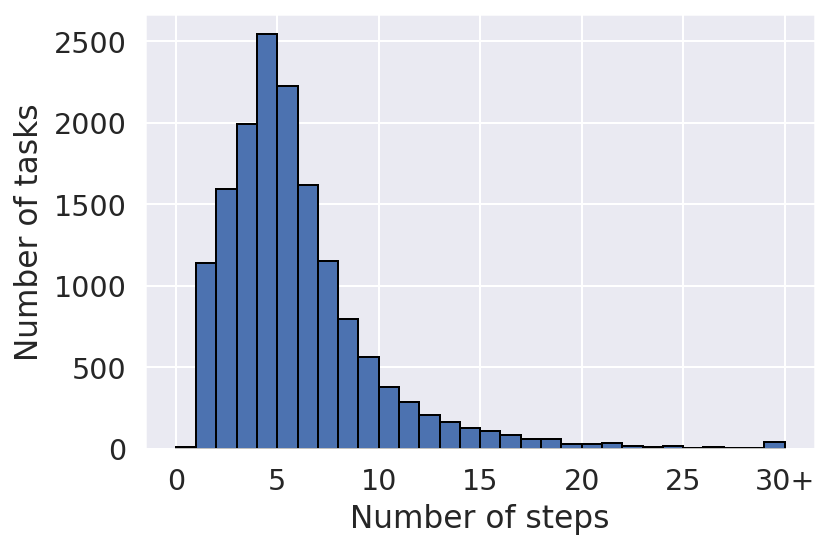}
    \vspace{-2.5ex}
    \caption{Task length distribution}
    \label{fig:task_length}
\end{subfigure}
\begin{subfigure}[t]{0.33\linewidth}
    \centering
    \includegraphics[width=\linewidth]{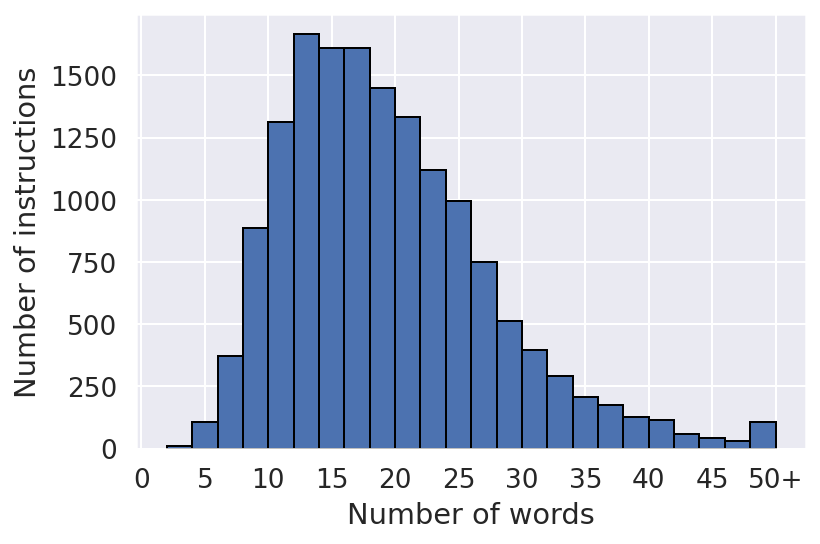}
    \vspace{-2.5ex}
    \caption{HL instr. length distribution}
    \label{fig:hl_words}
\end{subfigure}
\begin{subfigure}[t]{0.33\linewidth}
    \centering
\includegraphics[width=\linewidth]{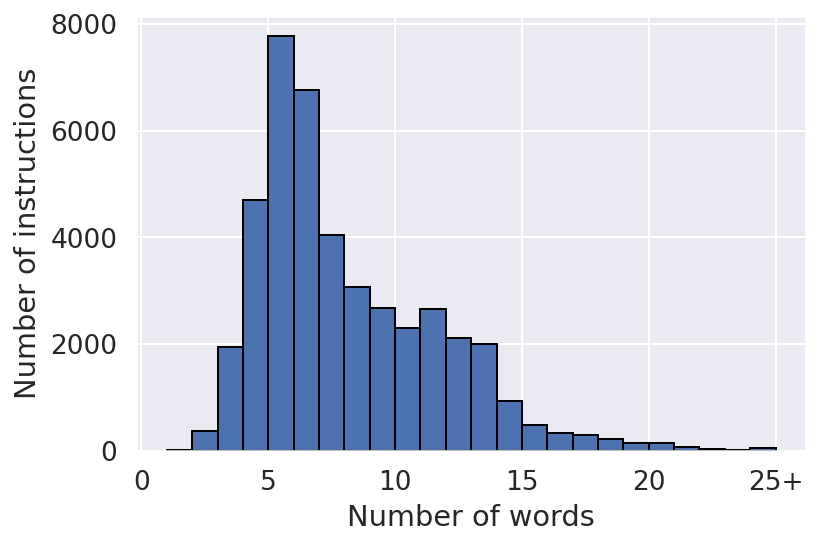}
    \vspace{-2.5ex}
    \caption{LL instr. length distribution}
    \label{fig:ll_words}
\end{subfigure}
\caption{Dataset statistics.}
\label{fig:dataset_summary}
\end{figure*}

\subsection{Dataset statistics}

Data statistics about \dataset are summarized in Table~\ref{table:dataset-comparison}. In addition, Figures~\ref{fig:elements} and \ref{fig:dataset_summary} report distributions of UI actions, task lengths, and lengths of high and low level instructions. The task length distribution (Figure~\ref{fig:task_length}), measured as number of steps required to complete the task, shows that tasks are of moderate length (between 1 and 13 steps for the 5th to 95th percentile, respectively). Lengths of high-level (HL) instructions fall between 8 and 34 words for the 5th and 95th percentile, and low-level (LL) instructions are between 3 and 14 words for the 5th and 95th percentile. 

%We found that the human-generated low-level instructions in our dataset were longer and more articulated than synthetically generated instructions in MoTIF \cite{motif}, which were between 3 and 11 words for the 5th and 95th percentiles.

\subsection{Dataset splits}

We create a train, a validation and 4 test splits whose number of task demonstrations (episodes) and characteristics are detailed in Table~\ref{table:splits-details}.
In order to measure how performance scales in domain and out of the domain of the collected data, we create the following test sub-splits: \emph{1) in domain data (IDD)}: randomly pulled episodes from the same distribution as the training data; \emph{2) app-unseen}: a test split using apps not present in the train split; \emph{3) task-unseen}: a test split with tasks not present in the train split; and \emph{4) category-unseen}: a test split with apps from categories not present in the train split.  Note that the test splits may contain overlapping episodes.  For example, episodes in the unseen-category split will also be in the unseen-app and unseen-tasks splits. It is also the case that the average number of elements per screen can vary significantly between training and testing. Appendix~\ref{sec:num_elements_in_test_splits} explains why that is the case.

\begin{table*}[t]
\centering
\caption{Details on \dataset train, validation and test splits. For every split, we report number of episodes, step actions, unique apps, app categories, and UI elements per screen.}
\vspace{-1ex}
\scalebox{0.86}{
\begin{tabular}{llcccccc}
\toprule
Split & Sub-splits & \# Episodes & \# Step actions & \# Apps & \# Categories &  Avg. \# elements per screen   \\
 \midrule
Train & - & 13,604~~~~ & 74,722~ & 769 & 39 &  222.2  \\ 
Val & - & 137 & ~~~690 & ~~99 & 29 & 214.4  \\ 
\midrule
\multirow{4}{*}{Test}  & IDD & 721 & 3,897 & 296 & 35  & 221.5   \\ 
& App-unseen & 631 & 3,475 & ~~64 & 12 & 185.4   \\ 
&Task-unseen & 803 & 4,464 & ~~90 & 12 & 181.6  \\ 
&Category-unseen & 700 & 3,891 & ~~68 & ~4  & 184.7 \\
 \bottomrule
\end{tabular}
}
\label{table:splits-details}
\end{table*}

\section{Experiments and results}

In order to test the impact of data scale and task complexity on transfer performance in domain and out of domain, we conduct experiments in which we train on different amounts of the data in the \dataset's training set.  We also test zero-shot and few-shot methods.

\subsection{Agent implementation}

%In a post-processing phase, we use the captured accessibility tree to extract several element metadata such as type (e.g., button, icon, etc.), textual information (e.g. accessibility labels, content information), and location bounds.
%Appendix~\ref{appendix:element-metadata} describes all element metadata we extract. \todo{complete appendix}

We implement a UI control agent for Android. The agent receives task instructions expressed in natural language. It observes the environment (the device) by deriving textual representations of the screen directly from the Android accessibility tree. The screen representation lists the on-screen UI elements. Each element is described according to the following attributes: type, text, content description, bounding boxes and various state tags (e.g., clickable, scrollable, focused, checked). As mobile UI screens may contain hundreds of UI elements (200 on average in \dataset, Table~\ref{table:splits-details}), we pre-process the screen representations to include only UI elements that have a non-empty text description or UI elements of critical types (switch and edit). This process facilitates the agent's task and reduces the input's size. %In our experiments we reduced the Android screen representations to be fewer than two thousand tokens on average; they are several thousands tokens only in extreme cases. 
Note that our agent implementation does not directly leverage the page screenshot. While recent work explores how to infer screen representations from raw screens~\cite{koh2024visualwebarena}, best performance is still reported when using accessibility trees or HTML~\cite{xie2024osworld,zheng2023seeact,android_world}. We expect the general trends we observe will hold true for multimodal language models.

During execution, the agent maintains a history over the previous steps. To avoid excessively large inputs, in the agent's input we include the screen description of only the current screen but append a history of the previously executed actions. In contrast to the action prediction output that locates UI elements by absolute coordinates, an action in the history is described in a self-contained manner, using its textual description and without any external reference. 

The agent predicts an action among a set of candidate actions. The set of available actions matches the actions defined by \dataset (Table~\ref{tab:action-space}) with two main changes. We add a \id{terminate} action that the agent predicts when it deems the task complete or infeasible. As this action is not originally provided in the dataset, for training purposes, we artificially insert it at the end of every episode (see Appendix~\ref{appendix:data-processing}). For efficiency reasons, as in prior work~\cite{uinav-sys}, the \id{input\_text} action is modified to include also the preceding click action necessary to set focus on the target element. The agent predicts the action type and any required arguments for the action, specifically the target UI element for a click action, the text to be typed and the target element for a typing action, the name of an app to open, the direction of a scroll, etc. For an example of screen representation, a summary of the agent's action space, and more details on the agent implementation please refer to Appendix~\ref{app:agent-impl}.

%We built a companion Android app utilizing the accessibility service. The companion app is responsible for sending accessibility tree as observation to the agent and executing agent predicted actions. Each agent action is executed as an macro actions~\cite{uinav-sys}. The execution of a macro is atomic as a new action can't start until the previous has finished. The companion app reports status of any macro action at the end of its execution, which is included in the history mentioned before.

%OR: as with the previous paragraph this is not relevant because we do offline evaluation
%Actions are executed on the actual phone or emulator using ADB (Android Debug Bridge). Each action type maps to specific ADB commands. For example, for click-based actions (click, long\_press) ADB simulates touch events at the specified coordinates on the screen. For typing actions, we first focus on the text input element and then use ADB to type the specified text; optionally the Enter button is clicked. Navigation actions (\id{navigate\_home}, \id{navigate\_back}) require sending corresponding key events to the device. To launch apps, ADB starts the target app.

\subsection{Experimental setup}
\label{sec:setup}

The LLMs we experiment with include PaLM-2L~\cite{palm2}, PaLM-2S~\cite{palm2}, Gemini 1.5 Pro~\cite{geminiteam2023gemini}, GPT-4 and GPT-4 Turbo~\cite{openai2024gpt4}. We set the temperature to zero for all models to obtain more deterministic responses. To limit compute, we perform fine-tuning only with PaLM-2S, and adopt the parameter efficient tuning approach of LoRA~\cite{LoRA}. We set the LoRA rank to 4 when fine-tuning with small amounts of data ($<1$k episodes), while switch to a rank of 64 when using more episodes. For few-shot experiments we use Gemini 1.5 Pro which provides a context window of 1M tokens. 

We create SeqIO~\cite{seqio} tasks to extract data points from \dataset and to generate prompts and target outputs (more details in Appendix~\ref{appendix:data-processing}). We setup two SeqIO tasks: (i) SeqIO HL (high-level) where only a high-level instruction is included in the prompt, and (ii) SeqIO LL (low-level) where both a low-level instruction and its corresponding high-level instruction are included. This second task emulates the use case where an LLM is used for decomposing high-level instructions into a sequence of low-level commands and another LLM is used for grounding; the assumption is that this grounding LLM may improve performance by having access to the context of the high-level command. In addition to the natural language instruction(s), each data point contains the textual description of the start screen, the history of performed actions, and the ground-truth action. Through these two SeqIO tasks, we investigate how a model performs on simpler (LL) or harder (HL) task instructions. 

To reduce LLM costs, some zero-shot and all few-shot evaluations are done on a subset of the test split of \dataset, \emph{Random-500}, that contains 500 random step actions from the full test split and has a similar sub-split distribution. We verified through experiments that results on Random-500 are a good approximation of the results on the full test split (Appendix~\ref{sec:random_500_vs_full_tests}).

\vspace{-0.5ex}
\paragraph{Zero-shot}
We test four zero-shot methods. (i) We use the AitW~\cite{aitw2023} prompt, specifically designed for Android and the PaLM model, without any modifications. (ii) We adapt the best-performing SeeAct~\cite{zheng2023seeact} variant ("choice") which grounds actions via textual choices. SeeAct was originally designed for GPT-4V for web navigation tasks. At each step, SeeAct queries the LLM twice. In the first query, it analyzes the state and performs reasoning. Then, in the second query, it asks the LLM to select an action from multiple choices. We use the SeeAct prompt by Rawles et al.~\cite{android_world} adapted to work on mobile and to take textual representations of Android screens as input. (iii) We evaluate the text-only version of M3A~\cite{android_world} that combines ReAct-style \cite{yao2022} and Reflexion-style \cite{reflexion} prompting.
(iv) Finally, we test a zero-shot prompt (implementation in Appendix~\ref{appendix:over-prompt}) of the same form we use with our agent described above.  This allows us to measure performance of a base model for our agent without any fine-tuning. This prompt emphasizes the use of a screen description composed of UI elements, hence the name \prompt (Element Representations). Note that with the exception of \prompt, which we ran with all 4 base models, to limit prompt sensitivity~\cite{sclar2023quantifying}, we ran the other prompts with the model family they were originally designed for.

\vspace{-0.5ex}
\paragraph{Few-shot and LoRA-tuned models}
When evaluating few-shot on HL instructions, samples drawn from the HL SeqIO task are used in the prompt. When testing on LL instructions, samples from the LL SeqIO task are included. For convenience, LoRA-tuned models are trained on a mixture of both the LL and HL SeqIO tasks as we found training on the two SeqIO tasks separately or in a mixture to achieve similar accuracy (see Appendix \ref{sec:level_of_instructions}). Best model checkpoints are selected using the validation split. 
We use the simple \prompt prompt for few-shot and LoRA-tuned models. 

% WB: I would remove the sentence below.  I don't think the a reviewer will accept this reasoning (a lot of papers have generated the needed examples for these methods - we could to), and at this point, I don't think a reader is necessarily going to be thinking about these alternate methods (so no need to put thoughts in their head).
%We cannot use prompts using reasoning such as chain of thought~\cite{wei2023chainofthought} or ReAct because they require collection or generation of additional training data.

%OR: too long and disclosing results we discuss later
%Particularly for fine-tuning, two factors are considered: 1) reasoning such as chain-of-thought or ReAct requires the collection or generation of additional data; 2) the base model for fine-tuning, PaLM 2S, lacks the reasoning capability of larger models. As a result, \prompt prompts a language model to output actions without any reasoning. Compared with SeeAct or M3A, \prompt is much shorter and simpler. But as shown in the next few sections, \prompt combined with fine-tuning reaches accuracy much higher than any one of the zero-shot and few-shot models, which is a proof that \prompt contains sufficient information for a grounder to work well. 

\vspace{-0.5ex}
\paragraph{Scaling analysis}
To conduct the scaling analysis, we vary the number of samples included in the prompt of few-shot (FS) techniques or in the training set of LoRA-tuned (LT) models. We randomly sample episodes from the SeqIO tasks using the following sample sizes: 5, 10, 100, 1k, 10k, and all (13,604) episodes. For few-shot only, to make the prompts more varied, we sample an equivalent number of step-examples from different episodes.

%, ``medium'': 10 episodes, ``large'': 100 episodes, ``xxl'': 1k episodes, ``xxl'': 10k episodes, and ``all'': 13,604 episodes.

%Few-shot takes a number of random samples of the same level of instructions as the testing task, while fine-tuning is provided with all the samples from a few random episodes. In order to boost the performance of few-shot, we pick the largest number of average per episode among all SeqIO tasks: 5.2 (Table~\ref{tbl:seqio-stats}), to decide the number of in-context samples equivalent to the number of episodes used for fine-tuning. 

\begin{table*}[t]
\centering
\caption{Performance on the IDD sub-split of Random-500. %For zero-shot we test the simple \prompt prompt as well as the AitW~\cite{aitw2023}, SeeAct~\cite{zheng2023seeact} and M3A~\cite{android_world} prompts. 
For few-shot (FS) and LoRA-tuned (LT) methods -$X$ (-$all$) indicates $X$ ($all$) episodes are used in the prompt or in training. Unless noted with a ``r64'' for models fine-tuned with a LoRA rank of 64, LoRA-tuned models use rank=4.}
\vspace{-1ex}
\scalebox{0.90}{
\begin{tabular}{lllcc}
\toprule
Regime & Method & Model &  \multicolumn{2}{c}{Step accuracy} \\
&  &           & high-level instr. & low-level instr. \\
 \midrule
\multirow{7}{*}{Zero-shot} & AitW & PaLM 2L   & 19.5 & \textbf{56.7} \\
 & SeeAct & GPT-4-Turbo  &    33.9 &  54.3  \\
 & M3A & GPT-4-Turbo  &    \textbf{42.1} &   55.0\\
 & \prompt & PaLM 2S  & 19.5 & 45.5 \\
 & \prompt & PaLM 2L & 33.0 & 45.9  \\
 & \prompt & GPT-4  &  32.1 & 51.7 \\
 & \prompt & Gemini 1.5 Pro  & 24.4 & 50.2 \\
\midrule
\multirow{3}{*}{Few-shot} & FS-5 & Gemini 1.5 Pro &  \textbf{41.8} & 50.2 \\ %26 examples
 & FS-10 & Gemini 1.5 Pro & 40.2 & 50.8 \\ %52 examples
 & FS-100 & Gemini 1.5 Pro & 39.5 & \textbf{53.3} \\ %520 examples
\midrule
\multirow{7}{*}{LoRA-tuned} & LT-5  & PaLM 2S   & 30.3 & 57.1 \\ %5->26 examples
 & LT-10 & PaLM 2S   & 28.5 & 58.9  \\ %10->52 examples
 & LT-100  & PaLM 2S  & 39.8 & 62.8 \\ %100->520 examples
 & LT-1k  & PaLM 2S  & 52.5 & 71.4   \\ %1k->5200 examples
 & LT-10k & PaLM 2S & 62.0 & 85.7 \\ %10->52000 examples
 & LT-all & PaLM 2S & 65.6 & 81.8  \\ %all examples
 & LT-1k-r64  & PaLM 2S  & 54.8  &  76.6  \\ 
 & LT-10k-r64 & PaLM 2S & 69.6 & 81.9 \\ %10->52000 examples
 & LT-all-r64 & PaLM 2S & \textbf{71.5} & \textbf{86.6} \\ %all examples
 \bottomrule
\end{tabular}
}
\label{tbl:grounding_perf}
\end{table*}

\vspace{-0.5ex}
\paragraph{Metrics}
As in prior work~\cite{zheng2023seeact,aitw2023}, as our evaluation metric we adopt step-wise accuracy, which measures the success of each task step. A step is successful if the predicted action and arguments (target element and text, if present) are correct. We adopt a relaxed metric that considers equivalent actions in additions to exact matches as successful (see Appendix~\ref{relaxed_action_matching} for details).

\subsection{In-domain performance}

%\todo{In domain performance: zero-shot < few-shot < fine-tuning on all training data.  Basically, this is the expected result - we don't spend much time here other than to point out that fine-tuning vastly outperforms the other two options so we set up the question: is fine-tuning a viable path forward? How much data would we need to get this to actually work.}

We start by evaluating zero-shot, few-shot and LoRA-tuned methods in domain. Table~\ref{tbl:grounding_perf} reports the step-wise accuracy performance on the IDD sub-split of Random-500. In-domain, LoRA-tuned models, despite using the smaller PaLM 2S model, when trained with sufficient amounts of data, largely outperform the zero-shot and few-shot methods. For low-level instructions, even LT-5 surpasses all non-fine-tuned models, while for high-level instructions, it requires more training data (1k episodes). The best fine-tuned model reaches 71.5\% on high-level instructions and 86.6\% on low-level instructions.

The best zero-shot performance on low-level instructions is obtained with AitW using PaLM 2L (56.7\%) and on high-level instructions with M3A using GPT-4 (42.1\%). This performance likely reflects the design of the prompts, the strength of the different base models used and the benefits of incorporating some high-level reasoning (included in M3A) for handling high-level instructions.  Interestingly, the few-shot performance is for the most part inferior to that of zero-shot methods.

%In fact, this prompt was specifically designed to handle low/medium level instructions which characterize the AitW dataset. When facing more complex, high-level instructions, AitW is much less accurate (19.5\%) which is lower than the very simple \prompt (33.0\%) using the same PaLM 2L model. M3A performs best than all other zero-shot prompts on high-level instructions. This is not surprising as this is the most advanced reasoning-based prompt we test.

%The few-shot performance slightly improves with more in-context exemplars on low-level instructions, but surpasses \prompt using the same Gemini model by only 3.1 percentage points. On high-level instructions, the few-shot performance slightly degrades when increasing the in-context exemplars, but it is more effective than fine-tuning when using the same number of samples (FS-5 achieves 41.8\% accuracy, while LT-5 achieves 30.3\% accuracy). 

\subsection{Effect of scale on in-domain transfer}

Fine-tuning obtains good performance in domain, but how much data is needed for acceptable performance? A failure at a single step may prevent task completion, a phenomenon we refer to as the ``weakest link effect.''   Making the simplifying assumption of i.i.d. success across steps, completing, for example, a 5-step task correctly 95\% of the time requires a 99\% step-wise accuracy.  We perform an analysis to extrapolate the amount of training data required to achieve such performance.

\begin{figure*}[t]
\centering
\begin{subfigure}[t]{0.47\textwidth}
    \includegraphics[width=\columnwidth]{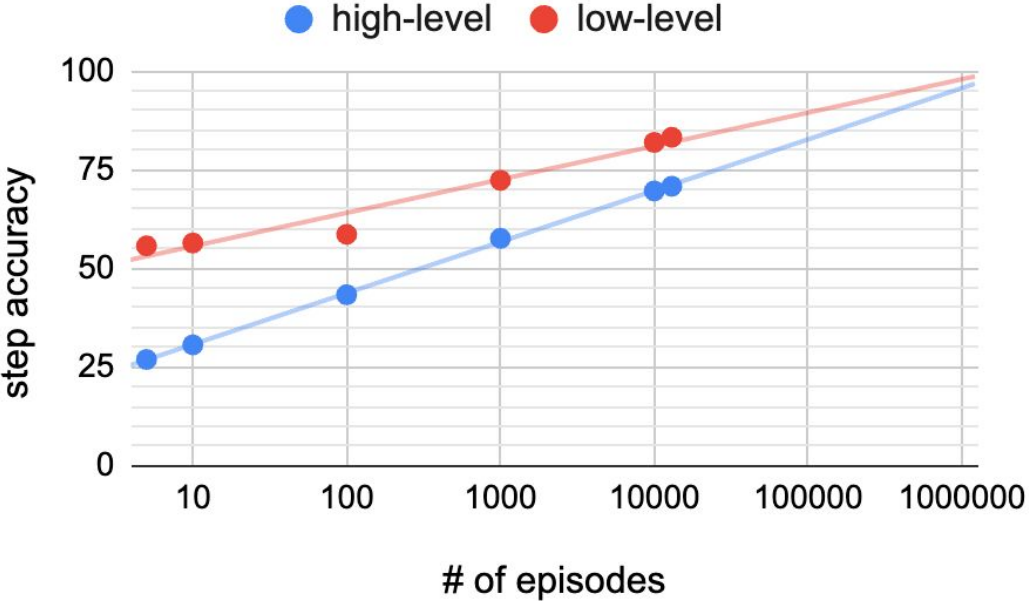}
    \caption{In-domain. High-level trendline $R^2=0.999$, low-level trendline $R^2=0.951$.}
    \label{fig:scaling-in-domain}
\end{subfigure}
\hfill
\begin{subfigure}[t]{0.47\textwidth}
    \includegraphics[width=\columnwidth]{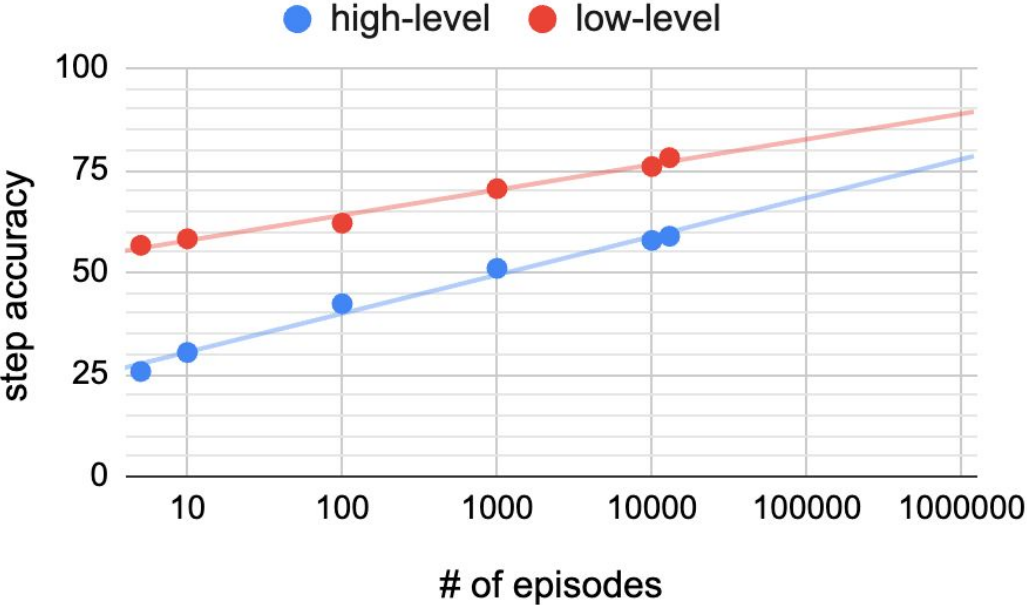}
    \caption{Out-of-domain. High-level trendline $R^2=0.986$, low-level trendline $R^2=0.987$.}
    \label{fig:scaling-out-domain}
\end{subfigure}
\caption{Relationship between step-wise accuracy and number of fine-tuning samples. With 5--100 episodes we use LoRA rank=4; with $\geq$1k episodes we use LoRA rank=64. IDD performance is based on the full IDD test split. OOD performance is based on the average across the three OOD splits.} 
\end{figure*}

Figure~\ref{fig:scaling-in-domain} visualizes the number of training episodes drawn from \dataset and the step accuracy achieved on the full IDD test sub-split. Both high and low-level curves exhibit linear trends ($R^2$ coefficients are greater than 0.95) with the log of training data. We extrapolate that it would take 500K and 1M episodes to reach $95\%$ step-accuracy for low and high-level instructions, respectively. However, while low-level tasks can be accomplished in one step, high-level tasks require multiple steps.  Taking 5 steps as the rough length for high-level tasks, we predict 2M episodes would be required to reach the 99\% step-wise accuracy to achieve 95\% episode completion for high-level tasks (see Appendix~\ref{sec:accuracy_vs_episode_length} for an empirical evaluation of performance as episode length is varied). While this is conservative by assuming no possible mistake recovery, we still feel this analysis provides a helpful rough quantification.

%By extrapolating to more training episodes, the IDD accuracy is estimated to reach 90\% and 98\% with 100K and 1M episodes, respectively, for low-level instructions, and 83\% (100K) and 95\% (1M) for high-level instructions.

\subsection{Effect of scale on out-of-domain transfer}

We now use \dataset's out-of-domain test splits (Table~\ref{table:splits-details}) to quantify how fine tuning with more demonstrations affects out-of-domain performance. This is important for assessing the robustness of agents used in the real world on tasks not foreseen in the data used to train an agent. %We use the \dataset's out-of-domain test splits (Table~\ref{table:splits-details}) for this analysis. %For zero-shot methods all test splits are effectively out of domain so in this comparison we use the best performing zero-shot tested on IDD (Table~\ref{tbl:grounding_perf}).

\begin{table}[t]
\centering
\caption{OOD step accuracy. In square brackets [X] we report the percentage point increase (+) or drop (-) from the IDD accuracy obtained on the full IDD test split.}
\vspace{-1ex}
\scalebox{0.87}{
\begin{tabular}{lcllll}
\toprule
 &  & IDD & app-unseen & task-unseen & categ-unseen \\
\midrule
\multirow{2}{*}{LT-5}     & HL & 26.9  & 25.7 \down{-1.2}  & 26.4 \down{-0.5}  & 25.1 \down{-1.8}  \\
                                &  LL & 55.7 & 56.9 \up{+1.2} & 56.6 \up{+0.9} & 56.4 \up{+0.7}\\
%\midrule
\multirow{2}{*}{LT-10}     & HL & 30.6  & 29.9 \down{-0.7}  & 31.1 \up{+0.5}  & 30.2 \down{-0.4}  \\
                                &  LL & 56.4 & 58.3 \up{+1.9} & 58.2 \up{+1.8}& 58.2 \up{+1.8}\\
%\midrule
\multirow{2}{*}{LT-100}      & HL & 43.3  & 42.4 \down{-0.9}  & 42.5 \down{-0.8}  & 42.1 \down{-1.2}  \\
                                &  LL & 58.6 & 62.7 \up{+4.1} & 61.7 \up{+3.1}& 61.8 \up{+3.2}\\
\midrule
\multirow{2}{*}{LT-1k}     & HL & 53.2  & 49.0 \down{-4.2}  & 49.3 \down{-3.9}  & 48.1 \down{-5.1}  \\
                                &   LL & 68.0 & 68.0 [~0.0]& 67.3 \down{-0.7} & 67.4 \down{-0.6}\\
\midrule
\multirow{2}{*}{LT-10k}  & HL & 63.9  & 55.2 \down{-8.7}  & 55.6 \down{-8.3}  & 54.2 \down{-7.7}  \\
                                &  LL & 78.7 & 76.7 \down{-2.0}& 75.6 \down{-3.1}& 75.5 \down{-3.2}\\
\midrule
\multirow{2}{*}{LT-all}     & HL & 65.5  & 58.7 \down{-6.8}  & 59.7 \down{-5.8}  & 58.2 \down{-7.3}  \\
                                &  LL & 80.7 & 78.6 \down{-2.1}& 77.9 \down{-2.8}& 77.8 \down{-2.9}\\
\midrule
\multirow{2}{*}{LT-1k-r64}     & HL &  57.6 & 51.1 \down{-6.5}  & 51.7 \down{-5.9}  & 50.2  \down{-7.4}  \\
                                &  LL & 72.3 & 71.0 \down{-1.3}& 70.4 \down{-1.9}& 70.1 \down{-2.2}\\
\midrule
\multirow{2}{*}{LT-10k-r64}     & HL & 69.6  & 57.7 \down{-11.9}  & 56.9 \down{-12.7}  & 58.9  \down{-10.7}  \\
                                &  LL & 81.9 & 76.3 \down{-5.6}& 75.8 \down{-6.1}& 75.2 \down{-6.7}\\
\midrule
\multirow{2}{*}{LT-all-r64}   & HL & 70.8  & 58.5 \down{-12.3}  & 59.6 \down{-11.2}  & 57.4 \down{-13.4}  \\
                                &  LL & 83.2 & 78.5 \down{-4.7}& 77.3 \down{-5.9}& 76.8 \down{-6.4}\\

 \bottomrule
\end{tabular}
}
%\vspace{-2ex}
\label{tbl:grounding_perf_by_subsplits}
\end{table}

As the number of fine-tuning samples increases, performance improves and so does the gap between IDD and OOD performance (Table~\ref{tbl:grounding_perf_by_subsplits}). With 10k or more episodes, the IDD accuracy is noticeably higher than on the three ODD splits. %The category-unseen split is the most challenging as it contains unseen apps and possibly also unseen tasks. For low-level instructions the gap increases more slowly. 
For example, with LT-10k (r=4), the gap is 7.7--8.7~pp for high-level instructions and 2.0--3.2~pp for low-level instructions. With r=64 the gap increases. In general, more out-of-domain transfer occurs for low-level tasks, which is expected as low-level tasks share more similarity across tasks and apps than high-level tasks.

As for in-domain, we extrapolate how much training data would be necessary to achieve a reasonable accuracy out of domain (Figure~\ref{fig:scaling-out-domain}). OOD step-accuracy grows more slowly than in domain, and is estimated to reach 95\% at 10M and 60M episodes for low-level and high-level instructions, respectively. Similar to above, the number of episodes we predict would be required to reach 99\% step accuracy to therefore achieve 95\% episode completion rate on 5-step high-level tasks is 150M. Based on these projections, it seems expensive but feasible to obtain good general LL performance with fine-tuning, while the predicted higher order of magnitude of the number of required demonstrations suggests fine-tuning alone may not be sufficient to achieve robust OOD performance on HL tasks.

\paragraph{More experiments.}
To complete our evaluation analysis in Appendix~\ref{app:more-results} we run more experiments studying the impact of other factors, including episode length, action types, and app types.

%\subsection{Ablations}
%\label{sec:ablations}

%We perform further ablations whose results are in reported in Appendix~\ref{app:ablations}. In short, we observe that as tasks get longer, the probability of predicting all individual steps correctly decreases such that the best performing LoRA-tuned model we presented (PaLM-2S-LT-all-r64) can complete 5-step tasks only in 21\% of the cases and 6-step tasks only in 7.6\% of the cases. For the same task lengths its step accuracy is 71.3\% and 64.1\%, respectively (Figure~\ref{fig:hl-ep-len}). As previously noted, this demonstrates that a step accuracy well above 90-95\% is necessary for real-world scenarios. Certain action predictions (e.g., predicting a task is completed) are particularly challenging (see analysis in Appendix~\ref{sec:confusion_matrix}), but anyway necessary for the whole task to succeed.

%We also observe the screen representations are critical to the success of UI control agents. We break down the performance of some of the tested models based on whether the tested apps are Google apps or non-Google apps. It is generally the case that Google proprietary apps (and high-trend apps) are more extensively annotated for accessibility, which leads UI control models to perform better on these apps, especially when given high-level instructions as input. As Table~\ref{tbl:first-party_vs_third-party} shows, on high-level instructions, the most best performing model (LT-all) achieves 82.5\% step accuracy on Google apps and only 58.7\% accuracy on non-Google apps.

\section{Limitations}
\label{sec:limitations}
There are multiple potential limitations in this work.  First, we only fine-tuned one model, PaLM-2S; however, while the absolute performance values would change, we expect our relative findings to be consistent across model families.  Additionally, using offline evaluation for agent performance has the known issue of not rewarding alternative routes to complete a task and the ability to take corrective actions~\cite{bishop2024latent,aitw2023}. Further, while selected to encompass important use cases, the set of app categories in \dataset is still an incomplete representation of all tasks users may ask agents to perform. Finally, studying inference costs is not a focus of the paper, although it is obvious that it is much cheaper to predict on a fine-tuned PaLM-2S model than on any larger models, such as PaLM-2L or GPT-4.

\section{Conclusion}
We have introduced \dataset, a large and diverse dataset structured for studying the performance of models in and out of domain on low and high-level tasks, as training data is scaled.  Using this dataset, we evaluate scaling of LoRA fine-tuned models.  We predict that to achieve 95\% accuracy for in-domain low-level tasks, 1M episodes would be required, while 2M episodes would be required to obtain 95\% episode completion rates for 5-step high-level tasks. While these results are for only one model, they suggest that fine-tuning may be a viable, though possibly expensive, route for obtaining high in-domain performance for low and high level tasks. Out of domain, 10M and 150M episodes would be required, respectively. This one to two orders of magnitude increase suggests fine-tuning may not scale well out of domain, and may not be sufficient to obtain good out-of-domain performance on HL tasks.  

%and out of domain performance on low-level tasks, while the order of magnitude more data required for OOD HL tasks, suggests 

%Motivated by the desire to evaluate to what extent scaling data size in fine-tuning can help us achieve real-world UI control agents, we collect and release \dataset. The dataset is large, diverse, and comprehensive, advancing existing datasets. We hope this dataset will stimulate further studies on generalizability of UI control agents in real world scenarios.

%\will{
%        \begin{itemize}
%            \item {There are some shortcomings reviewers might point out: what if we fine-tuned larger models? Why didn't we use multimodal? We should address those head on.}
%            \item{We should bring home conclusion: grounding may be achievable via fine-tuning! This is huge.  Grounding is half the battle.}
%            \item{On the other hand, we can argue we don't see a path-forward for high-level tasks that can be achieved via fine-tuning alone (I think we want to be careful to say fine-tuning will be part of a successful approach, but we will also need more.)  We should suggest what those other ingredients might be (tool use, etc...).}
%            \item{Should point out that we obtained these results with small models - sure, maybe big models would give better performance after fine-tuning, but we still showed there is a viable path forward with small models and that is exciting for thinking about real applications where latency matters.}
%        \end{itemize}
%  }      
        
%\section*{Acknowledgements}
%The authors thank...

% \bibliographystyle{abbrv}
\bibliographystyle{plain}
\bibliography{custom}

\begin{thebibliography}{10}

\bibitem{andreassen2021evolution}
Anders Andreassen, Yasaman Bahri, Behnam Neyshabur, and Rebecca Roelofs.
\newblock The evolution of out-of-distribution robustness throughout
  fine-tuning.
\newblock {\em arXiv preprint arXiv:2106.15831}, 2021.

\bibitem{bai2021uibert}
Chongyang Bai, Xiaoxue Zang, Ying Xu, Srinivas Sunkara, Abhinav Rastogi,
  Jindong Chen, and Blaise~Ag{\"{u}}era y~Arcas.
\newblock {UIBert}: Learning generic multimodal representations for {UI}
  understanding.
\newblock In Zhi{-}Hua Zhou, editor, {\em Proc. of the 30th International Joint
  Conference on Artificial Intelligence, {IJCAI} 2021}, pages 1705--1712.
  ijcai.org, 2021.

\bibitem{bishop2024latent}
William~E Bishop, Alice Li, Christopher Rawles, and Oriana Riva.
\newblock Latent state estimation helps ui agents to reason, 2024.

\bibitem{bonatti2024windowsagentarenaevaluating}
Rogerio Bonatti, Dan Zhao, Francesco Bonacci, Dillon Dupont, Sara Abdali,
  Yinheng Li, Yadong Lu, Justin Wagle, Kazuhito Koishida, Arthur Bucker,
  Lawrence Jang, and Zack Hui.
\newblock {Windows Agent Arena: Evaluating Multi-Modal OS Agents at Scale},
  2024.

\bibitem{gpt3}
Tom~B. Brown, Benjamin Mann, Nick Ryder, Melanie Subbiah, Jared Kaplan,
  Prafulla Dhariwal, Arvind Neelakantan, Pranav Shyam, Girish Sastry, Amanda
  Askell, Sandhini Agarwal, Ariel Herbert-Voss, Gretchen Krueger, Tom Henighan,
  Rewon Child, Aditya Ramesh, Daniel~M. Ziegler, Jeffrey Wu, Clemens Winter,
  Christopher Hesse, Mark Chen, Eric Sigler, Mateusz Litwin, Scott Gray,
  Benjamin Chess, Jack Clark, Christopher Berner, Sam McCandlish, Alec Radford,
  Ilya Sutskever, and Dario Amodei.
\newblock Language models are few-shot learners, 2020.

\bibitem{motif}
Andrea Burns, Deniz Arsan, Sanjna Agrawal, Ranjitha Kumar, Kate Saenko, and
  Bryan~A. Plummer.
\newblock Mobile app tasks with iterative feedback {(MoTIF)}: Addressing task
  feasibility in interactive visual environments.
\newblock {\em CoRR}, abs/2104.08560, 2021.

\bibitem{deng2023mind2web}
Xiang Deng, Yu~Gu, Boyuan Zheng, Shijie Chen, Samuel Stevens, Boshi Wang, Huan
  Sun, and Yu~Su.
\newblock {Mind2Web}: Towards a generalist agent for the web, 2023.

\bibitem{geminiteam2023gemini}
{Gemini Team}.
\newblock Gemini: A family of highly capable multimodal models, 2023.

\bibitem{palm2}
{Google}, Rohan Anil, Andrew~M. Dai, Orhan Firat, Melvin Johnson, Dmitry
  Lepikhin, Alexandre Passos, Siamak Shakeri, Emanuel Taropa, Paige Bailey,
  Zhifeng Chen, Eric Chu, Jonathan~H. Clark, Laurent~El Shafey, Yanping Huang,
  Kathy Meier-Hellstern, Gaurav Mishra, Erica Moreira, Mark Omernick, Kevin
  Robinson, Sebastian Ruder, Yi~Tay, Kefan Xiao, Yuanzhong Xu, Yujing Zhang,
  Gustavo~Hernandez Abrego, Junwhan Ahn, Jacob Austin, Paul Barham, Jan Botha,
  James Bradbury, Siddhartha Brahma, Kevin Brooks, Michele Catasta, Yong Cheng,
  Colin Cherry, Christopher~A. Choquette-Choo, Aakanksha Chowdhery, Clément
  Crepy, Shachi Dave, Mostafa Dehghani, Sunipa Dev, Jacob Devlin, Mark Díaz,
  Nan Du, Ethan Dyer, Vlad Feinberg, Fangxiaoyu Feng, Vlad Fienber, Markus
  Freitag, Xavier Garcia, Sebastian Gehrmann, Lucas Gonzalez, Guy Gur-Ari,
  Steven Hand, Hadi Hashemi, Le~Hou, Joshua Howland, Andrea Hu, Jeffrey Hui,
  Jeremy Hurwitz, Michael Isard, Abe Ittycheriah, Matthew Jagielski, Wenhao
  Jia, Kathleen Kenealy, Maxim Krikun, Sneha Kudugunta, Chang Lan, Katherine
  Lee, Benjamin Lee, Eric Li, Music Li, Wei Li, YaGuang Li, Jian Li, Hyeontaek
  Lim, Hanzhao Lin, Zhongtao Liu, Frederick Liu, Marcello Maggioni, Aroma
  Mahendru, Joshua Maynez, Vedant Misra, Maysam Moussalem, Zachary Nado, John
  Nham, Eric Ni, Andrew Nystrom, Alicia Parrish, Marie Pellat, Martin Polacek,
  Alex Polozov, Reiner Pope, Siyuan Qiao, Emily Reif, Bryan Richter, Parker
  Riley, Alex~Castro Ros, Aurko Roy, Brennan Saeta, Rajkumar Samuel, Renee
  Shelby, Ambrose Slone, Daniel Smilkov, David~R. So, Daniel Sohn, Simon
  Tokumine, Dasha Valter, Vijay Vasudevan, Kiran Vodrahalli, Xuezhi Wang,
  Pidong Wang, Zirui Wang, Tao Wang, John Wieting, Yuhuai Wu, Kelvin Xu, Yunhan
  Xu, Linting Xue, Pengcheng Yin, Jiahui Yu, Qiao Zhang, Steven Zheng,
  Ce~Zheng, Weikang Zhou, Denny Zhou, Slav Petrov, and Yonghui Wu.
\newblock {PaLM 2} technical report, 2023.

\bibitem{webagent:iclr2024}
Izzeddin Gur, Hiroki Furuta, Austin Huang, Mustafa Safdari, Yutaka Matsuo,
  Douglas Eck, and Aleksandra Faust.
\newblock A real-world webagent with planning, long context understanding, and
  program synthesis.
\newblock In {\em The 12th International Conference on Learning
  Representations}, 2024.

\bibitem{learning-navigate-web}
Izzeddin Gur, Ulrich Rueckert, Aleksandra Faust, and Dilek Hakkani-Tur.
\newblock {Learning to Navigate the Web}.
\newblock In {\em 7th International Conference on Learning Representations
  (ICLR '19)}, May 6--9 2019.

\bibitem{he2024webvoyager}
Hongliang He, Wenlin Yao, Kaixin Ma, Wenhao Yu, Yong Dai, Hongming Zhang,
  Zhenzhong Lan, and Dong Yu.
\newblock {WebVoyager}: Building an end-to-end web agent with large multimodal
  models, 2024.

\bibitem{hernandez2021scaling}
Danny Hernandez, Jared Kaplan, Tom Henighan, and Sam McCandlish.
\newblock Scaling laws for transfer, 2021.

\bibitem{LoRA}
Edward~J Hu, Yelong Shen, Phillip Wallis, Zeyuan Allen-Zhu, Yuanzhi Li, Shean
  Wang, Lu~Wang, and Weizhu Chen.
\newblock {LoRA}: {Low-Rank} adaptation of large language models, June 2021.

\bibitem{pmlr-v162-humphreys22a}
Peter~C Humphreys, David Raposo, Tobias Pohlen, Gregory Thornton, Rachita
  Chhaparia, Alistair Muldal, Josh Abramson, Petko Georgiev, Adam Santoro, and
  Timothy Lillicrap.
\newblock A data-driven approach for learning to control computers.
\newblock In Kamalika Chaudhuri, Stefanie Jegelka, Le~Song, Csaba Szepesvari,
  Gang Niu, and Sivan Sabato, editors, {\em Proceedings of the 39th
  International Conference on Machine Learning}, volume 162 of {\em Proceedings
  of Machine Learning Research}, pages 9466--9482. PMLR, 17--23 Jul 2022.

\bibitem{kaplan2020scaling}
Jared Kaplan, Sam McCandlish, Tom Henighan, Tom~B. Brown, Benjamin Chess, Rewon
  Child, Scott Gray, Alec Radford, Jeffrey Wu, and Dario Amodei.
\newblock Scaling laws for neural language models, 2020.

\bibitem{rci_kim2023language}
Geunwoo Kim, Pierre Baldi, and Stephen~Marcus McAleer.
\newblock Language models can solve computer tasks.
\newblock In {\em 37th Conference on Neural Information Processing Systems},
  2023.

\bibitem{koh2024visualwebarena}
Jing~Yu Koh, Robert Lo, Lawrence Jang, Vikram Duvvur, Ming~Chong Lim, Po-Yu
  Huang, Graham Neubig, Shuyan Zhou, Ruslan Salakhutdinov, and Daniel Fried.
\newblock {VisualWebArena}: Evaluating multimodal agents on realistic visual
  web tasks, 2024.

\bibitem{kojima-nips22}
Takeshi Kojima, Shixiang~(Shane) Gu, Machel Reid, Yutaka Matsuo, and Yusuke
  Iwasawa.
\newblock Large language models are zero-shot reasoners.
\newblock In S.~Koyejo, S.~Mohamed, A.~Agarwal, D.~Belgrave, K.~Cho, and A.~Oh,
  editors, {\em Advances in Neural Information Processing Systems}, volume~35,
  pages 22199--22213. Curran Associates, Inc., 2022.

\bibitem{kumar2022finetuning}
Ananya Kumar, Aditi Raghunathan, Robbie~Matthew Jones, Tengyu Ma, and Percy
  Liang.
\newblock Fine-tuning can distort pretrained features and underperform
  out-of-distribution.
\newblock In {\em International Conference on Learning Representations}, 2022.

\bibitem{uinav-sys}
Wei Li, Fu-Lin Hsu, Will Bishop, Folawiyo Campbell-Ajala, Max Lin, and Oriana
  Riva.
\newblock {UINav}: A practical approach to train on-device automation agents.
\newblock In {\em North American Chapter of the Association for Computational
  Linguistics}, 2023.

\bibitem{li-acl20}
Yang Li, Jiacong He, Xin Zhou, Yuan Zhang, and Jason Baldridge.
\newblock Mapping natural language instructions to mobile {UI} action
  sequences.
\newblock In {\em Proc. of the 58th Annual Meeting of the Association for
  Computational Linguistics, {ACL} 2020, Online, July 5-10, 2020}, pages
  8198--8210. Association for Computational Linguistics, 2020.

\bibitem{web-workflows18}
Evan~Zheran Liu, Kelvin Guu, Panupong Pasupat, and Percy Liang.
\newblock Reinforcement learning on web interfaces using workflow-guided
  exploration.
\newblock In {\em 6th International Conference on Learning Representations
  (ICLR '18)}, 2018.

\bibitem{lu2024weblinx}
Xing~Han Lù, Zdeněk Kasner, and Siva Reddy.
\newblock {WebLINX}: Real-world website navigation with multi-turn dialogue,
  2024.

\bibitem{nakano21:webgpt}
Reiichiro Nakano, Jacob Hilton, Suchir Balaji, Jeff Wu, Long Ouyang, Christina
  Kim, Christopher Hesse, Shantanu Jain, Vineet Kosaraju, William Saunders,
  Xu~Jiang, Karl Cobbe, Tyna Eloundou, Gretchen Krueger, Kevin Button, Matthew
  Knight, Benjamin Chess, and John Schulman.
\newblock {WebGPT}: Browser-assisted question-answering with human feedback,
  2021.

\bibitem{openai2024gpt4}
OpenAI.
\newblock Gpt-4 technical report, 2024.

\bibitem{aitw2023}
Chris Rawles, Alice Li, Daniel Rodriguez, Oriana Riva, and Timothy Lillicrap.
\newblock {Android in the Wild}: A large-scale dataset for android device
  control.
\newblock In {\em NeurIPS 2023 Datasets and Benchmarks Track}, 2023.

\bibitem{android_world}
Christopher Rawles, Sarah Clinckemaillie, Yifan Chang, Jonathan Waltz,
  Gabrielle Lau, Marybeth Fair, Alice Li, William Bishop, Wei Li, Folawiyo
  Campbell-Ajala, Daniel Toyama, Robert Berry, Divya Tyamagundlu, Timothy
  Lillicrap, and Oriana Riva.
\newblock {AndroidWorld}: A dynamic benchmarking environment for autonomous
  agents, May 2024.

\bibitem{seqio}
Adam Roberts, Hyung~Won Chung, Anselm Levskaya, Gaurav Mishra, James Bradbury,
  Daniel Andor, Sharan Narang, Brian Lester, Colin Gaffney, Afroz Mohiuddin,
  Curtis Hawthorne, Aitor Lewkowycz, Alex Salcianu, Marc van Zee, Jacob Austin,
  Sebastian Goodman, Livio~Baldini Soares, Haitang Hu, Sasha Tsvyashchenko,
  Aakanksha Chowdhery, Jasmijn Bastings, Jannis Bulian, Xavier Garcia, Jianmo
  Ni, Andrew Chen, Kathleen Kenealy, Jonathan~H. Clark, Stephan Lee, Dan
  Garrette, James Lee-Thorp, Colin Raffel, Noam Shazeer, Marvin Ritter, Maarten
  Bosma, Alexandre Passos, Jeremy Maitin-Shepard, Noah Fiedel, Mark Omernick,
  Brennan Saeta, Ryan Sepassi, Alexander Spiridonov, Joshua Newlan, and Andrea
  Gesmundo.
\newblock Scaling up models and data with $\texttt{t5x}$ and $\texttt{seqio}$.
\newblock {\em arXiv preprint arXiv:2203.17189}, 2022.

\bibitem{sclar2023quantifying}
Melanie Sclar, Yejin Choi, Yulia Tsvetkov, and Alane Suhr.
\newblock Quantifying language models' sensitivity to spurious features in
  prompt design or: How i learned to start worrying about prompt formatting,
  2023.

\bibitem{miniwob}
Tianlin Shi, Andrej Karpathy, Linxi Fan, Jonathan Hernandez, and Percy Liang.
\newblock World of bits: An open-domain platform for web-based agents.
\newblock In Doina Precup and Yee~Whye Teh, editors, {\em Proc. of the 34th
  International Conference on Machine Learning}, volume~70 of {\em Proceedings
  of Machine Learning Research}, pages 3135--3144. PMLR, 06--11 Aug 2017.

\bibitem{reflexion}
Noah Shinn, Federico Cassano, Edward Berman, Ashwin Gopinath, Karthik
  Narasimhan, and Shunyu Yao.
\newblock Reflexion: Language agents with verbal reinforcement learning, March
  2023.

\bibitem{ugif}
Sagar~Gubbi Venkatesh, Partha Talukdar, and Srini Narayanan.
\newblock {UGIF}: Ui grounded instruction following, 2022.

\bibitem{wang2022self}
Xuezhi Wang, Jason Wei, Dale Schuurmans, Quoc Le, Ed~Chi, Sharan Narang,
  Aakanksha Chowdhery, and Denny Zhou.
\newblock Self-consistency improves chain of thought reasoning in language
  models.
\newblock {\em arXiv preprint arXiv:2203.11171}, 2022.

\bibitem{wei2021finetuned}
Jason Wei, Maarten Bosma, Vincent Zhao, Kelvin Guu, Adams~Wei Yu, Brian Lester,
  Nan Du, Andrew~M Dai, and Quoc~V Le.
\newblock Finetuned language models are zero-shot learners.
\newblock In {\em International Conference on Learning Representations}, 2021.

\bibitem{wei2023chainofthought}
Jason Wei, Xuezhi Wang, Dale Schuurmans, Maarten Bosma, Brian Ichter, Fei Xia,
  Ed~Chi, Quoc Le, and Denny Zhou.
\newblock Chain-of-thought prompting elicits reasoning in large language
  models, 2023.

\bibitem{pmlr-v162-wortsman22a}
Mitchell Wortsman, Gabriel Ilharco, Samir~Ya Gadre, Rebecca Roelofs, Raphael
  Gontijo-Lopes, Ari~S Morcos, Hongseok Namkoong, Ali Farhadi, Yair Carmon,
  Simon Kornblith, and Ludwig Schmidt.
\newblock Model soups: averaging weights of multiple fine-tuned models improves
  accuracy without increasing inference time.
\newblock In Kamalika Chaudhuri, Stefanie Jegelka, Le~Song, Csaba Szepesvari,
  Gang Niu, and Sivan Sabato, editors, {\em Proceedings of the 39th
  International Conference on Machine Learning}, volume 162 of {\em Proceedings
  of Machine Learning Research}, pages 23965--23998. PMLR, 17--23 Jul 2022.

\bibitem{xie2024osworld}
Tianbao Xie, Danyang Zhang, Jixuan Chen, Xiaochuan Li, Siheng Zhao, Ruisheng
  Cao, Toh~Jing Hua, Zhoujun Cheng, Dongchan Shin, Fangyu Lei, Yitao Liu,
  Yiheng Xu, Shuyan Zhou, Silvio Savarese, Caiming Xiong, Victor Zhong, and Tao
  Yu.
\newblock {OSWorld}: Benchmarking multimodal agents for open-ended tasks in
  real computer environments, 2024.

\bibitem{yan2023gpt4v}
An~Yan, Zhengyuan Yang, Wanrong Zhu, Kevin Lin, Linjie Li, Jianfeng Wang,
  Jianwei Yang, Yiwu Zhong, Julian McAuley, Jianfeng Gao, Zicheng Liu, and
  Lijuan Wang.
\newblock {GPT-4V in Wonderland}: Large multimodal models for zero-shot
  smartphone {GUI} navigation, 2023.

\bibitem{yao2023webshop}
Shunyu Yao, Howard Chen, John Yang, and Karthik Narasimhan.
\newblock {WebShop}: Towards scalable real-world web interaction with grounded
  language agents, 2023.

\bibitem{yao2023}
Shunyu Yao, Dian Yu, Jeffrey Zhao, Izhak Shafran, Thomas~L Griffiths, Yuan Cao,
  and Karthik Narasimhan.
\newblock Tree of thoughts: Deliberate problem solving with large language
  models.
\newblock {\em arXiv preprint arXiv:2305.10601}, 2023.

\bibitem{yao2022}
Shunyu Yao, Jeffrey Zhao, Dian Yu, Nan Du, Izhak Shafran, Karthik~R Narasimhan,
  and Yuan Cao.
\newblock {ReAct}: Synergizing reasoning and acting in language models.
\newblock In {\em The 11th International Conference on Learning Representations
  (ICLR)}, 2023.

\bibitem{zheng2023seeact}
Boyuan Zheng, Boyu Gou, Jihyung Kil, Huan Sun, and Yu~Su.
\newblock {GPT-4V(ision)} is a generalist web agent, if grounded.
\newblock {\em arXiv preprint arXiv:2401.01614}, 2024.

\bibitem{zheng2023synapse}
Longtao Zheng, Rundong Wang, Xinrun Wang, and Bo~An.
\newblock Synapse: Trajectory-as-exemplar prompting with memory for computer
  control.
\newblock In {\em The 12 International Conference on Learning Representations
  (ICLR'24)}, 2024.

\bibitem{zhou2022least}
Denny Zhou, Nathanael Sch{\"a}rli, Le~Hou, Jason Wei, Nathan Scales, Xuezhi
  Wang, Dale Schuurmans, Claire Cui, Olivier Bousquet, Quoc Le, et~al.
\newblock Least-to-most prompting enables complex reasoning in large language
  models.
\newblock {\em arXiv preprint arXiv:2205.10625}, 2022.

\bibitem{zhou2024webarena}
Shuyan Zhou, Frank~F. Xu, Hao Zhu, Xuhui Zhou, Robert Lo, Abishek Sridhar,
  Xianyi Cheng, Tianyue Ou, Yonatan Bisk, Daniel Fried, Uri Alon, and Graham
  Neubig.
\newblock {WebArena}: A realistic web environment for building autonomous
  agents.
\newblock In {\em Proc. of the 12th International Conference on Learning
  Representations}, 2024.

\end{thebibliography}

\newpage
\begin{appendix}

\section{Ethical considerations}
\label{sec:potential_negative_societal_impacts}

%Autonomous UI agents if exploited maliciously may complete harmful tasks. An agent may also unintentionally perform actions due to model mistakes which in the worst case causes safety concerns. 

Autonomous UI control agents can bring value to visually-impaired users, by providing them with access to a much wider range of applications and functionality. More broadly, they can enhance human productivity by automating everyday tasks. UI agents have societal, security and privacy implications. An agent may leak private information or carry out a task in an unacceptable way or produce unwanted side effects. Malicious actors could also use these agents for undesired purposes such as overriding anti-fraud mechanisms or manipulating applications to achieve undesirable goals. For these reasons, deployment of this technology going forward will have to be carefully considered and combined with research in other areas on LLM safety to balance potential societal trade-offs with risks.

In our experiments, we used the PaLM 2 model, which is available publicly through the Vertex AI PaLM API from Google. Our research use was in accordance with Google's AI prohibited use policy (\url{https://policies.google.com/terms/generative-ai/use-policy}).

\section{Dataset details}
\label{app:dataset}

\subsection{Data collection}
\label{app:data-collection}

The data collection was carried out by annotators who are paid contractors, who received a standard contracted wage, which complies with living wage laws in their country of employment. The annotators were informed of the intended use of the data collected and signed a data usage agreement. They did not use their personal devices nor they were required to enter any private information.

We provided annotators with a detailed instructional document and video tutorials on how to operate the Android and web apps for data collection. All raters went through a training phase where they could familiarize with the tools and received personalized feedback based on manual inspection of the collected traces.  

\begin{figure*}
\makebox[\textwidth][c]{\includegraphics[width=1.5\textwidth]{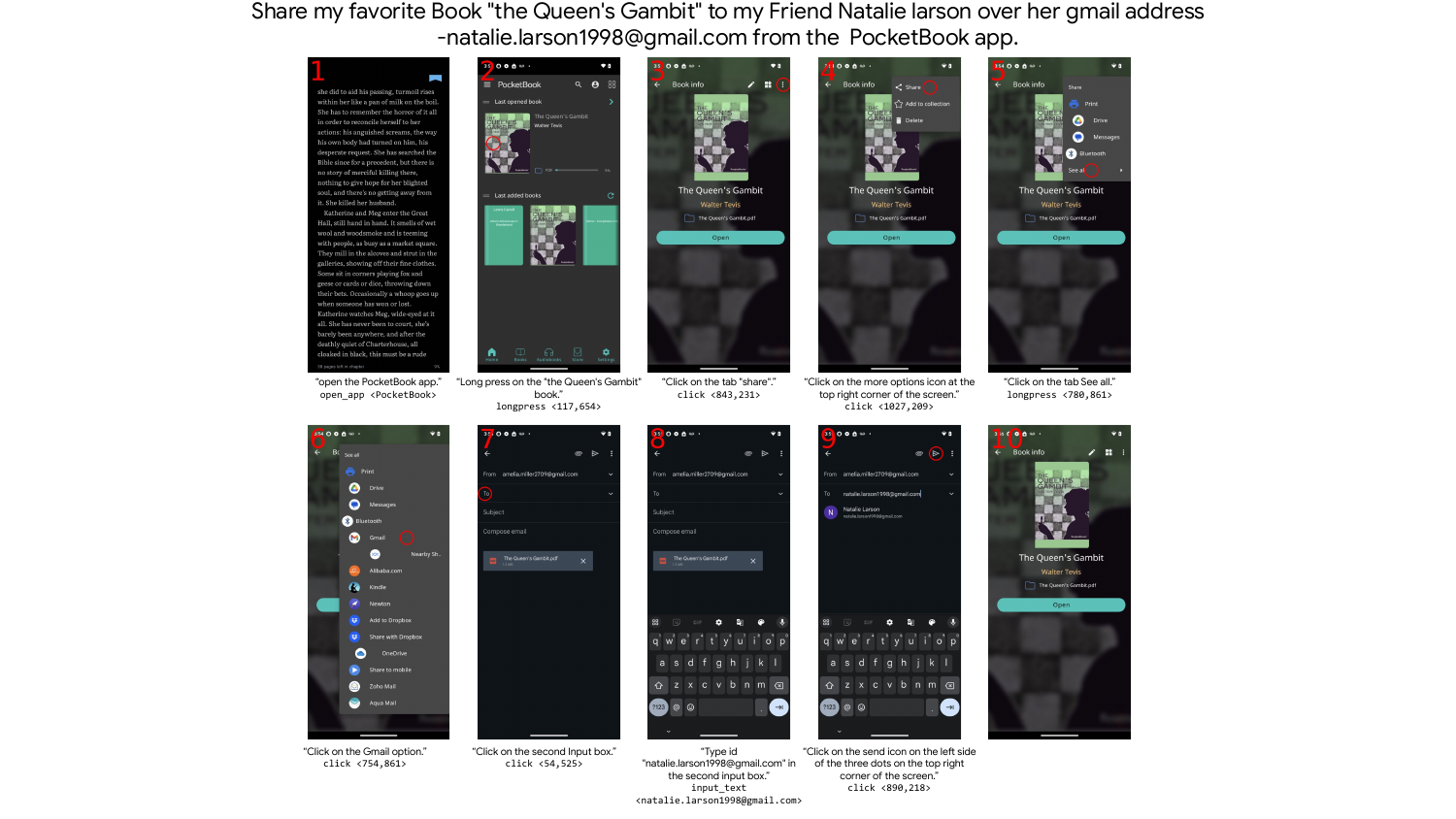}}
\makebox[\textwidth][c]{\includegraphics[width=1.4\textwidth]{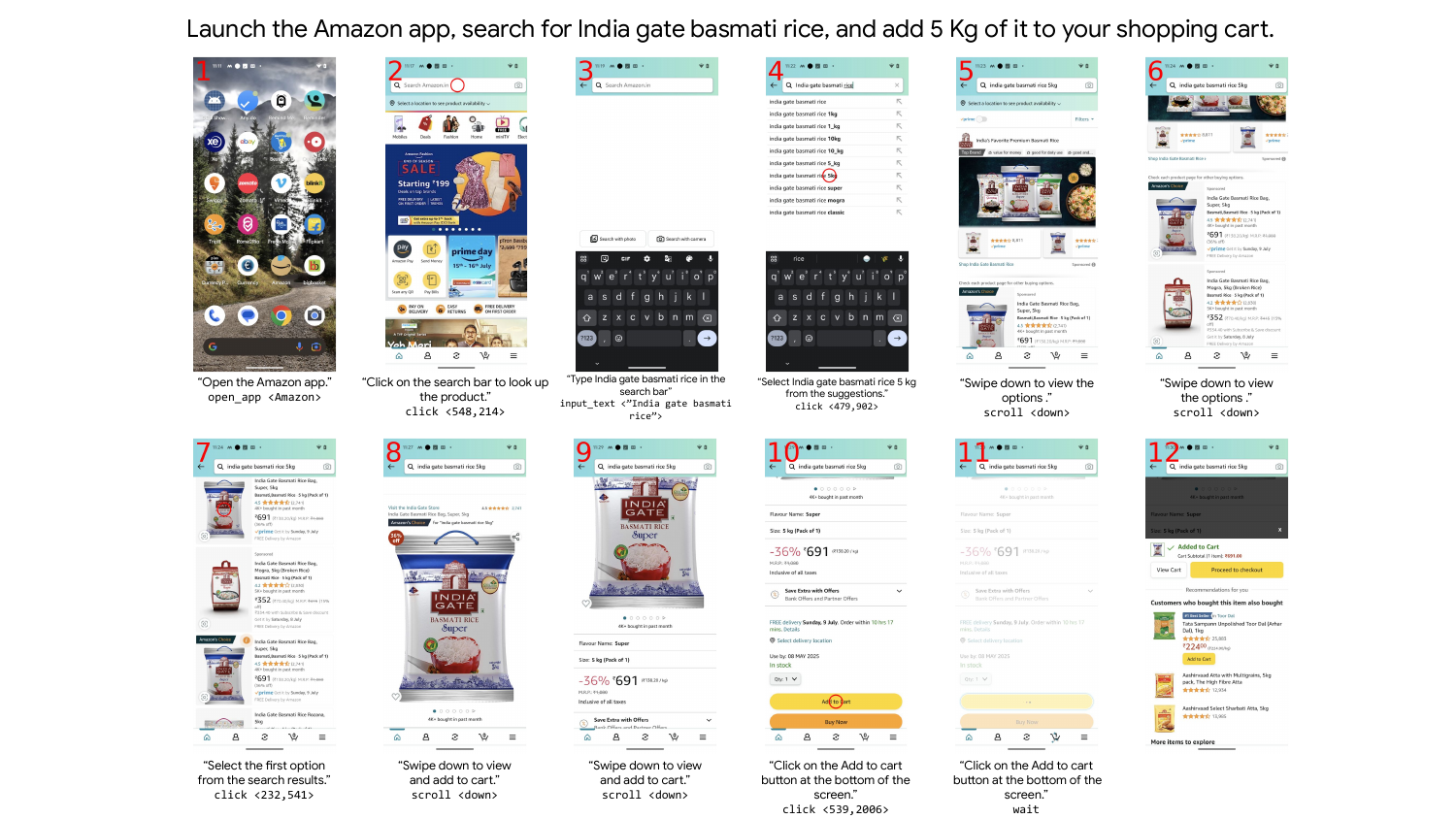}}
%\vspace{-2ex}
\caption{Example episodes contained in \dataset. Red circles highlight the location of click and long press actions on the screen. Red numbers are added only for illustration purposes.}
\label{fig:example-traces}
\end{figure*}

Examples of episodes from \dataset are shown in Figure~\ref{fig:example-traces}.  % \todo{wei?alice?}

\subsection{Dataset format}
\label{app:data-format}

%https://g3doc.corp.google.com/third_party/google_research/google_research/android_control/README.md?cl=637079336

\dataset is publicly released at \url{https://github.com/google-research/google-research/tree/master/android_control}. Each datapoint is stored as a TFRecord file with the following fields:

\begin{itemize}
    \item \emph{episode\_id}: a unique identifier integer for each episode. This is especially useful when generating the data splits.
    \item \emph{goal}: the high-level instruction for the entire episode.
    \item \emph{screenshots}: a list of screenshot byte strings for each observation encoded as PNGs.
    \item \emph{accessibility\_trees}: a list of Android accessibility trees for each observation.
    \item \emph{screenshot\_widths}: a list of the widths of each of the screenshots.
    \item \emph{screenshot\_heights}: a list of the heights of each of the screenshots.
    \item \emph{actions}: a list of actions represented as JSON dictionaries. The actions are performed between consecutive screenshots, so there are len(screenshots) - 1 of them.
    \item \emph{step\_instructions}: a list of the low-level instructions describing each step to complete the task. The number of step instructions equals the number of actions, but it is important to note that each step instruction does not necessarily describe a single action. A step instruction can require more than one action to complete, and in these cases the step instruction is repeated to maintain a one-to-one mapping from step instructions to actions.
\end{itemize}

\subsection{Accessibility node metadata}
\label{appendix:element-metadata}

Each node in the Android accessibility tree corresponds to a UI element in the screen. Each node is described by multiple metadata. In our UI control agent implementation we use the following node metadata:

\begin{itemize}
    \item Element type: \id{class\_name} (e.g. \id{Button}, \id{TextView}, \id{Image}, etc.)
    \item Textual attributes: \id{text}, \id{content\_description}, \id{hint\_text}, \id{tooltip\_text}, \id{view\_id\_resource\_name}.  
    \item Location and size: \id{bounds\_in\_screen}.
    \item Element status (as boolean): \id{is\_checked}, \id{is\_enabled}, \id{is\_focused}, \id{is\_selected}.
    \item Element properties (as boolean): \id{is\_checkable}, \id{is\_clickable}, \id{is\_editable},  \id{is\_focusable}, \id{is\_long\_clickable}, \id{is\_scrollable},  \id{is\_password}, \id{is\_visible\_to\_user}.
\end{itemize}

\subsection{Number of UI elements in test splits}
\label{sec:num_elements_in_test_splits}

\begin{figure*}[h]
\centering
\includegraphics[width=\columnwidth]{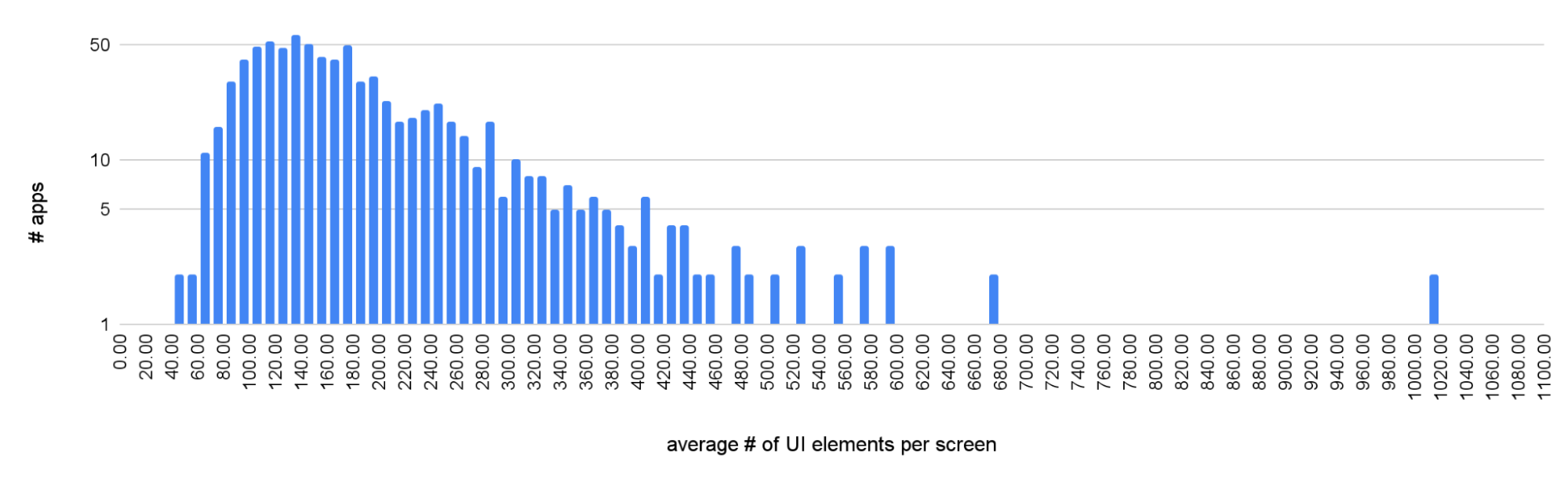}
 \vspace{-1ex}
\caption{Histogram of average number of UI elements per app for all apps in the \dataset dataset. Please note the y-axis is logarithmic.} 
\label{fig:num_ui_elements_per_screen}
\end{figure*}

In Table~\ref{table:splits-details}, we note that in the test sub-split that is sampled from the same distribution as the train split (the “IDD” split), the average number of UI elements per screen is comparable, as expected, while in the other 3 out-of-domain test sub-splits there is a significant drop in the number of UI elements per screen. To explain this phenomenon, we calculated the average number of UI elements across all screens in the data for each app (for simplicity we call this “the average number of UI elements per app”). Figure~\ref{fig:num_ui_elements_per_screen} shows the distribution of the average number of UI elements per app for the entire dataset. The distribution shows a long tail, with a few apps with a very large number of UI elements per screen. If we remove these outliers by dropping the top 5\% of apps with the most elements per screen, the average number of UI elements across the remaining apps is 180.9, which is inline with the out-of-domain test sub-splits. A reasonable question is then why do the out-of-domain test sub-splits not sample from the tail? We believe this is because, as shown in Table~\ref{table:splits-details}, the number of apps in each sub-split is relatively small. Further, there is overlap between the different test sub-splits (this was done to keep the total size of the test set reasonable, if each sub-split needed to be disjoint, it would leave less train data), so the total number of apps across the union of all out-of-domain test sub-splits is still modest, again supporting that we simply did not draw enough outliner apps from the long tail to affect the average for the out-of-domain splits.

\section{UI control agent implementation}
\label{app:agent-impl}

% \todo{here goes anything that is specific to our agent implementation, e.g., screen representations, action definitions and additional actions such as terminate, data pre-processing, etc.}

\subsection{Observation space} 
\label{sec:observation}

The device state is perceived through the UI screen currently displayed. A screen representation is derived from the Android accessibility tree and lists all UI elements composing the UI. Each element is described by three fields: text (a textual description derived from the element's textual description, content description, class name, etc.), position, and status (e.g., whether a checkbox is selected). These fields are populated using the metadata (or a combination thereof) associated with Android accessibility nodes (Appendix~\ref{appendix:element-metadata}). For simplicity, in this paper, we only experiment with screen descriptions that consist of a flat list of UI elements. Figure~\ref{fig:example_screenshot} shows an example screenshot and the corresponding JSON screen representation.

%OR: as with the previous paragraph this is not relevant because we do offline evaluation
%Actions are executed on the actual phone or emulator using ADB (Android Debug Bridge). Each action type maps to specific ADB commands. For example, for click-based actions (click, long\_press) ADB simulates touch events at the specified coordinates on the screen. For typing actions, we first focus on the text input element and then use ADB to type the specified text; optionally the Enter button is clicked. Navigation actions (\id{navigate\_home}, \id{navigate\_back}) require sending corresponding key events to the device. To launch apps, ADB starts the target app.

\subsection{Action space}
\label{sec:json_actions}

When predicting an action that involves a target element, the model should output sufficient details to locate the target UI element, either an index or its geometric information such as its bounding rectangle. To reduce the complexity of parsing the model output, we prompt an LLM to output its action selection in a predefined JSON format. In the case of element click, for instance, the model outputs a  prediction in the following format: \id{\{"action\_type":"click","x":<x\_coordinate>,"y":<y\_coordinate>\}}, where the target element is identified by its center coordinates. We found LLMs work equally well with predicting element centers or element indices, but as the former approach is compatible with click actions that are not restricted to specific UI elements, our implementation always outputs the center of the target UI element. The same applies to all actions that take an element as input.

Table~\ref{tbl:json_actions} lists the JSON action templates and defines the agent's action space. Compared to the actions collected in \dataset (Table~\ref{tab:action-space}), there are two main differences. First, we introduce the \id{type} action derived from the dataset's \id{input\_text} action. This action is obtained by aggregating the \id{input\_text} action and its preceeding \id{click} action, which is necessary to focus on the UI element before typing. Accordingly, the low-level task instructions are merged via concatenation. This unified action is more efficient from an agent implementation's perspective and reduces latency at execution time. Second, we introduce a new action, \id{terminate}, which signals whether the agent deems the task as successfully completed or infeasible. To support training and testing of this action, we insert an additional step at the end of each episode with low-level instruction ``terminate', \id{action=terminate}, and value set to ``successful'' or ``infeasible'' depending on the episode status.

%has the option of either predicting the index of the target element as \id{\{"action\_type":"click","index":<element\_index>\}}, or the center coordinate of the target element as \id{\{"action\_type":"click","x":<x\_coordinate>,"y":<y\_coordinate>\}}. 

%We found that LLMs work equally well either predicting element indices or element centers. As clicking coordinate is compatible with click actions that are not restricted to UI elements, in all our experiments, a model always outputs the center of the target UI element when referencing UI elements. The same applies to \id{long\_press} and \id{input\_text} actions.

%As the input signals of determining the end condition of grounding are the same for selecting actions, we combine the two into a single model. The grounder outputs one of two status actions when it thinks a grounding is successful or infeasible. 

\begin{table}[]
\centering
\caption{Agent's action space. For each action the table reports the JSON template the agent is asked to predict.}
\vspace{-1ex}
\scalebox{0.85}{
\begin{tabular}{ll}
\toprule
 Action &  JSON template \\
 \midrule
 click & \{"action\_type":"click","x":<x\_coordinate>,"y":<y\_coordinate>\}\\
 long\_press & \{"action\_type":"long\_press","x":<x\_coordinate>,"y":<y\_coordinate>\}\\
 type & \{"action\_type":"type","text":<text\_input>,"x":<x\_coordinate>,"y":<y\_coordinate>\}\\
 scroll & \{"action\_type":"scroll","direction":<up, down, left, or right>\}\\
 \midrule
 navigate\_home &\{"action\_type":"navigate\_home"\}\\
 navigate\_back &\{"action\_type":"navigate\_back"\}\\
 open\_app &\{"action\_type":"open\_app","app\_name":<app\_name>\}\\
 wait & \{"action\_type":"wait"\}\\
 \midrule
 terminate & \{"action\_type":"status","goal\_status":<"successful","infeasible">\}\\
 \bottomrule
\end{tabular}
}
\label{tbl:json_actions}
\end{table}

\subsection{History}
Actions performed in previous steps and their outcome are included as the history of the current step. An action in history is derived from the JSON action so that it is self-contained without any external reference. Section~\ref{sec:example_screenshot_action_history} contains an example of history. Please note that in an offline dataset, as in this paper, the outcome of a previous action is recorded by annotators and most likely successful. However, in a real system, the underlying framework can report an action as failed due to screen synchronization errors or prediction errors which may render a target element not localizable or an action not executable. 

%An action in a history is self-contained without any external reference. For example an element click action is represented as \id{click~[element\_text]}, where the xy coordinates of the center of the element is replaced by its text label. \S\ref{sec:example_screenshot_action_history} contains a real example of history.    

\subsection{Examples of screen description, JSON action, and history} \label{sec:example_screenshot_action_history}

\begin{figure}[h]
  \begin{subfigure}[]{.65\linewidth}
  \centering
\begin{lstlisting}[escapeinside={(*}{*)}]
{"UI elements":[{"center":[350,1601],"size":[323,77],"text":"Clear history"},{"center":[128,1491],"size":[131,60],"text":"Display"},{"center":[158,1420],"size":[190,81],"text":"Display"},{"center":[320,1265],"size":[514,59],"text":"Location > Location services"},{"center":[317,1194],"size":[509,82],"text":"Bluetooth scanning"},{"center":[128,1038],"size":[131,60],"text":"Display"},{"center":[271,967],"size":[416,81],"text":"Brightness level"},{"center":[943,772],"size":[188,225],(*\textcolor{red}{"text":"Switch"}*),(*\textcolor{blue}{"checked":true}*)},{"center":[497,812],"size":[616,60],"text":"Connected devices > Connection"},{"center":[317,741],"size":[256,81],"text":"Bluetooth"},{"center":[94,772],"size":[63,225],"text":""},{"center":[943,546],"size":[188,225],(*\textcolor{red}{"text":"Switch"}*),(*\textcolor{blue}{"checked":true}*)},{"center":[449,586],"size":[520,59],"text":"Network & internet > Internet"},{"center":[257,516],"size":[136,81],"text":"Wi-Fi"},{"center":[94,546],"size":[63,225],"text":""},{"center":[566,370],"size":[1027,126],"text":"RECENT SEARCH RESULTS"},{"center":[603,219],"size":[953,126],"text":"Search settings"},{"center":[63,219],"size":[126,147],"text":"Back"},{"center":[996,68],"size":[19,136],"text":"Battery 100 percent."},{"center":[950,67],"size":[38,38],"text":""},{"center":[892,67],"size":[78,41],"text":"No internet"},{"center":[348,68],"size":[57,136],"text":"Android System notification: C"},{"center":[290,68],"size":[58,136],"text":"Android System notification: S"},{"center":[209,68],"size":[105,136],"text":"09:22"}]}
\end{lstlisting}
  \end{subfigure}
  \hfill
  \begin{subfigure}[]{0.27\linewidth}
    \centering
    \includegraphics[width=1.1\textwidth]{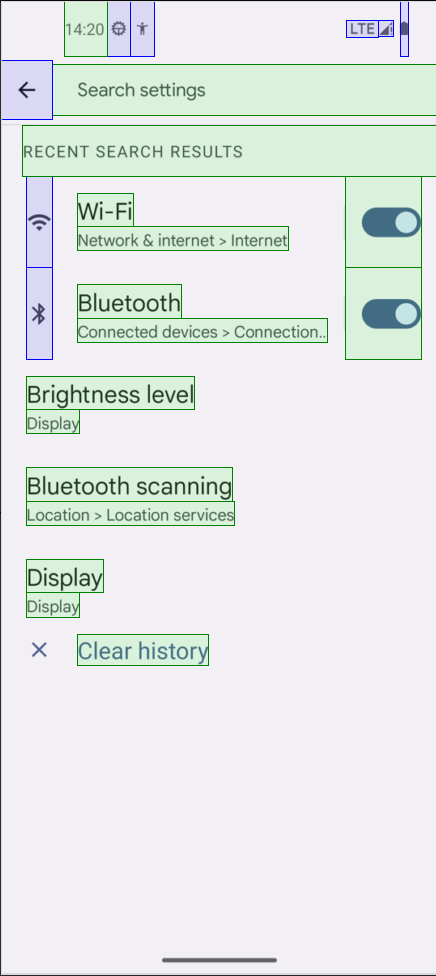}
    %\caption{An example screenshot annotated with UI elements. Note that this example is picked purposely to have few UI elements so that the screen description is compact.}
  \end{subfigure}
  \caption{An example screenshot annotated with UI elements (right) and corresponding screen representation (left). Note that this example is picked purposely to have few UI elements so that the screen description is compact.}
  \label{fig:example_screenshot}
\end{figure}

Figure~\ref{fig:example_screenshot} shows an example screenshot annotated with UI elements, and the list next to it is the corresponding screen description. The position and shape of each UI element are defined by "center" and "size" while its semantic meaning is described by the "text" field. Note that a switch element does not have any textual attribute, therefore a text label "Switch" derived from its \id{class\_name} is assigned (text in red), and its status is specified by the "checked" field (text in blue). 

%\begin{wrapfigure}{R}{0.32\textwidth}  % 'r' for right side, 0.5\textwidth for half the text width
%    \centering
%    \includegraphics[width=0.24\textwidth]{images/screen_shot_example.png}
%    \caption{An example screenshot annotated with UI elements. Note that this example is picked purposely to have few UI elements so that the screen description is compact.}
%    \label{fig:example_screenshot}
%\end{wrapfigure}

%\begin{lstlisting}
%{"UI elements":[{"center":[950,2078],"size":[132,172],"text":"Google Lens"},{"center":[818,2078],"size":[131,172],"text":"Voice search"},{"center":[539,2078],"size":[976,172],"text":"Search"},{"center":[953,1845],"size":[205,168],"text":"Camera"},{"center":[747,1845],"size":[206,168],"text":"Chrome"},{"center":[540,1845],"size":[207,168],"text":"Play Store"},{"center":[333,1845],"size":[206,168],"text":"Messages"},{"center":[126,1845],"size":[207,168],"text":"Phone"},{"center":[561,253],"size":[989,72],"text":"Wed, May 22"},{"center":[540,1148],"size":[1080,2131],"text":"Home"},{"center":[992,34],"size":[21,62],"text":"Battery 100 percent."},{"center":[944,34],"size":[41,60],"text":"Phone signal full."},{"center":[899,34],"size":[47,60],"text":"Wifi signal full."},{"center":[219,34],"size":[60,62],"text":"Settings notification: Virtual"},{"center":[132,34],"size":[111,62],"text":"19:40"}]}
%\end{lstlisting}

Given a goal example : "search for lord of the rings". The ground truth output is an action in JSON format,
%\begin{lstlisting}
\id{{"action\_type":"click","x":539,"y":2078}}
%\end{lstlisting}
, where (539, 2078) is the center of the search bar at the bottom.

The following is an example of action history that is included in the prompt after two actions:

\begin{lstlisting}
{"0":["click [Search]","successful"], "1":["type "lord of the rings" at [Search apps, web and more]","successful"]}
\end{lstlisting}

Note that target elements, such as [Search] and [Search apps, web and more], are identified by their text labels or descriptions, hence do not reference the corresponding screen description.

\section{Experimental details}\label{sec:experimental_details}

\subsection{Data processing and training details}
\label{appendix:data-processing}

To run our experiments we generate 2 SeqIO tasks (HL and LL) that process the \dataset dataset as follows. (i) In order to support prediction of task completion actions not present in the original dataset, at the end of every episode we artificially insert a \id{terminate} action that takes the episode status (successful or infeasible) as argument. (ii) The dataset omits identifiers for target elements associated with \id{click} or \id{long\_press} actions. Such identifiers can be retrieved by determining which accessibility node encompasses the click location. When multiple nodes satisfy this criterion, the node with the smallest area is selected. However, we discard any step with an element-based action (\id{click}, \id{long\_press}) that does not have a UI element associated. This is due to either a touch on an empty area or a target UI element missing from the accessibility tree. These discarded steps are still considered in the action history to support prediction of later steps that reference previous actions or elements. (iii) Finally, we discard (only from SeqIO LL tasks) steps that are missing a low-level instruction which annotators may have forgotten to enter. %The discarded steps account for roughly 10\% of all steps in the full dataset, while the equivalent number of episodes in the SeqIO tasks are not significantly affected as 
An episode is dropped only if all of its steps are discarded. Table~\ref{tbl:seqio-stats} shows the statistics of the SeqIO tasks that are the results of this processing.  

\begin{table*}[t]
\centering
\caption{Statistics of the SeqIO tasks generated from \dataset. Note that the number of steps here reported is different than what reported in Table~\ref{table:splits-details} because it accounts for \id{terminate} and \id{type} actions which are not present in the original dataset (see \ref{sec:json_actions}).}
\vspace{-1ex}
\scalebox{0.89}{
\begin{tabular}{llcccccc}
\toprule
Split & SeqIO tasks & \# Episodes & \# Steps & Avg \# steps per episode   \\
 \midrule
\multirow{2}{*}{Train} & HL & 13,604 & 70,796 & 5.2\\ 
                       & LL & 13,344 & 60,954 & 4.6\\
\midrule                       
\multirow{2}{*}{Validation} & HL & ~~~~~137 & ~~~~~655 & 4.8 \\ 
                       & LL & ~~~~135 & ~~~~559 & 4.1\\
\midrule                       
\multirow{2}{*}{Test} & HL & ~~7,897 & ~~1,543  & 5.1 \\ 
                       & LL & ~~6,833 & ~~1,500  & 4.6\\
 \bottomrule
\end{tabular}
}
\label{tbl:seqio-stats}
\end{table*}

\subsection{LLM prompts for zero-shot experiments}
\label{appendix:prompts}

For an LLM-based UI control agent, its prompt describes what the agent is expected to do, the action space and the expected format of actions, the task instruction, the current screen and application, and the history of previously executed actions and their outcome.

In our experiments we test four different prompts. We use the original AitW and M3A prompt as described in the original publications~\cite{aitw2023,android_world}. For the SeeAct~\cite{zheng2023seeact} and \prompt prompts we list the detailed prompts in the following. Please note that variable placeholders that are replaced by real values are marked by \{\{\{ \}\}\}. 

\subsubsection{SeeAct prompt}
\label{appendix:seeact-prompt}

We took the Android adaptation of the SeeAct~\cite{zheng2023seeact} prompt by Rawles et al.~\cite{android_world}. As it is designed for GPT-4V, we slightly modify it by replacing its annotated pixel input by using the same JSON screen description as shown in Section~\ref{sec:observation}. We find that by using our element filtering approach fewer than 50 elements are usually present per screen. As a result, the ranker model of SeeAct which selects candidate UI elements is not necessary, hence it is disabled.

\exclude{
We slightly modify the original SeeAct prompt designed for GPT-4V, by replacing its web input with a textual screen description derived from Android accessibility trees. We find that by using our element filtering approach (see Appendix~\ref{sec:observation}) fewer than 50 elements are usually present per screen. As a result, the ranker model of SeeAct which selects candidate UI elements is not necessary, hence it is disabled. Additionally, we augment the original SeeAct's action space to support actions specific to mobile: swipe, long-press, navigate home, navigate back and launch apps. The modified prompt is as follows.

The prompt for analyzing the state:
\begin{lstlisting}
Imagine that you are imitating humans operating an Android device for a task step by step. At each stage, you can see the Android screen like humans by a screenshot and know the previous actions before the current step decided by yourself through recorded history. You need to decide on the first following action to take. You can tap on an element, long-press an element, swipe, input text, open an app, or use the keyboard enter, home, or back key. (For your understanding, they are like `adb shell input tap`, `adb shell input swipe`, `adb shell input text`, `adb shell am start -n`, and `adb shell input keyevent`). One next step means one operation within these actions. Unlike humans, for typing (e.g., in text areas, text boxes), you should try directly typing the input or selecting the choice, bypassing the need for an initial click. You should not attempt to create accounts, log in or do the final submission. Terminate when you deem the task complete or if it requires potentially harmful actions.You are asked to complete the following task: {{{grounding_goal}}}

Previous Actions:
{{{previous_actions}}}

The screenshot below shows the Android screen you see. Follow the following guidance to think step by step before outlining the next action step at the current stage:

(Current Webpage Identification)
Firstly, think about what the current screen is.

(Previous Action Analysis)
Secondly, combined with the screenshot, analyze each step of the previous action history and their intention one by one. Particularly, pay more attention to the last step, which may be more related to what you should do now as the next step. Specifically, if the last action involved a INPUT TEXT, always evaluate whether it necessitates a confirmation step, because typically a single INPUT TEXT action does not make effect. (often, simply pressing 'Enter', assuming the default element involved in the last action, unless other clear elements are present for operation).

(Screenshot Details Analysis)
Closely examine the screenshot to check the status of every part of the screen to understand what you can operate with and what has been set or completed. You should closely examine the screenshot details to see what steps have been completed by previous actions even though you are given the textual previous actions. Because the textual history may not clearly and sufficiently record some effects of previous actions, you should closely evaluate the status of every part of the webpage to understand what you have done.

(Next Action Based on Android screen and Analysis)
Then, based on your analysis, in conjunction with human phone operation habits and the logic of app design, decide on the following action. And clearly outline which element on the Android screen users will operate with as the first next target element, its detailed location, and the corresponding operation.

To be successful, it is important to follow the following rules:
1. You should only issue a valid action given the current observation.
2. You should only issue one action at a time
3. For handling the select dropdown elements on a screen, it's not necessary for you to provide completely accurate options right now. The full list of options for these elements will be supplied later.

The following screen description in JSON represents the key information of the screenshot. It is composed of a list of UI elements with each UI element depicted by its attributes.
# Explanation of inputs
The top edge of a screen has y_coordinate equal to 0. The y_coordinate of the bottom edge of a screen equals to screen height. In screen_description, missing the 'checked' field for an element indicates that it is NOT checked.
The size of an element is defined by width and height.

# Screen description
{{{screen_description}}}
Start your analysis from here:
\end{lstlisting}

The prompt for selecting an action from multiple choices:
\begin{lstlisting}
Imagine that you are imitating humans operating an Android device for a task step by step. At each stage, you can see the Android screen like humans by a screenshot and know the previous actions before the current step decided by yourself through recorded history. You need to decide on the first following action to take. You can tap on an element, long-press an element, swipe, input text, open an app, or use the keyboard enter, home, or back key. (For your understanding, they are like `adb shell input tap`, `adb shell input swipe`, `adb shell input text`, `adb shell am start -n`, and `adb shell input keyevent`). One next step means one operation within these actions. Unlike humans, for typing (e.g., in text areas, text boxes), you should try directly typing the input or selecting the choice, bypassing the need for an initial click. You should not attempt to create accounts, log in or do the final submission. Terminate when you deem the task complete or if it requires potentially harmful actions.You are asked to complete the following task: {{{grounding_goal}}}

Previous Actions:

The screenshot below shows the Android screen you see. Follow the following guidance to think step by step before outlining the next action step at the current stage:

(Current Webpage Identification)
Firstly, think about what the current screen is.

(Previous Action Analysis)
Secondly, combined with the screenshot, analyze each step of the previous action history and their intention one by one. Particularly, pay more attention to the last step, which may be more related to what you should do now as the next step. Specifically, if the last action involved a INPUT TEXT, always evaluate whether it necessitates a confirmation step, because typically a single INPUT TEXT action does not make effect. (often, simply pressing 'Enter', assuming the default element involved in the last action, unless other clear elements are present for operation).

(Screenshot Details Analysis)
Closely examine the screenshot to check the status of every part of the screen to understand what you can operate with and what has been set or completed. You should closely examine the screenshot details to see what steps have been completed by previous actions even though you are given the textual previous actions. Because the textual history may not clearly and sufficiently record some effects of previous actions, you should closely evaluate the status of every part of the webpage to understand what you have done.

(Next Action Based on Android screen and Analysis)
Then, based on your analysis, in conjunction with human phone operation habits and the logic of app design, decide on the following action. And clearly outline which element on the Android screen users will operate with as the first next target element, its detailed location, and the corresponding operation.

To be successful, it is important to follow the following rules:
1. You should only issue a valid action given the current observation.
2. You should only issue one action at a time
3. For handling the select dropdown elements on a screen, it's not necessary for you to provide completely accurate options right now. The full list of options for these elements will be supplied later.

The following screen description in JSON represents the key information of the screenshot. It is composed of a list of UI elements with each UI element depicted by its attributes.
# Explanation of inputs
The top edge of a screen has y_coordinate equal to 0. The y_coordinate of the bottom edge of a screen equals to screen height. In screen_description, missing the 'checked' field for an element indicates that it is NOT checked.
The size of an element is defined by width and height.

# Screen description
{{{screen_description}}}

(Reiteration)
First, reiterate your next target element, its detailed location, and the corresponding operation.

(Multichoice Question)
Below is a multi-choice question, where the choices are elements in the webpage. All elements are arranged in the order based on their height on the webpage, from top to bottom (and from left to right). This arrangement can be used to locate them. From the screenshot, find out where and what each one is on the webpage, taking into account both their text content and HTML details. Then, determine whether one matches your target element. Please examine the choices one by one. Choose the matching one. If multiple options match your answer, choose the most likely one by re-examining the screenshot, the choices, and your further reasoning.

{{{multiple_choices}}}
If none of these elements match your target element, please select AL. None of the other options match the correct element.

(Final Answer)
Finally, conclude your answer using the format below. Ensure your answer is strictly adhering to the format provided below. Please do not leave any explanation in your answers of the final standardized format part, and this final part should be clear and certain. The element choice, action, and value should be in three separate lines.

Format:

ELEMENT: The uppercase letter of your choice. (No need for {NAVIGATE HOME, KEYBOARD ENTER, TERMINATE, SWIPE, NAVIGATE BACK, WAIT, OPEN APP, ANSWER})

ACTION: Choose an action from {NAVIGATE HOME, KEYBOARD ENTER, INPUT TEXT, SWIPE, TERMINATE, LONG PRESS, CLICK, NAVIGATE BACK, WAIT, OPEN APP, ANSWER}.

VALUE: Provide additional input based on ACTION.

The VALUE means:
If ACTION == INPUT TEXT, specify the text to be typed.
If ACTION == SWIPE, specify the direction: up, down, left, right.
If ACTION == OPEN APP, provide the name of the app to be opened.
If ACTION == ANSWER, specify the text of your answer to respond directly to a question or request for information.
For CLICK, LONG PRESS, KEYBOARD ENTER, NAVIGATE HOME, NAVIGATE BACK, WAIT, and TERMINATE, write "None".
\end{lstlisting}
}
\subsubsection{\prompt prompt}
\label{appendix:over-prompt}

We also design a new prompt, \prompt, which emphasizes the use of a screen description composed of UI elements. The prompt does not encourage an LLM to reason, hence it is much simpler than the other prompts. This prompt is also used for few-shot and fine-tuning experiments. The prompt is shown below:

\begin{lstlisting}
An agent follows instructions on an Android device. Each instruction requires one or more steps. At each step, the input includes previous_actions, active_app, screen_width_and_height, and screen_description. You are required to select one action from the available actions.

# Available actions:
{"action_type":"click","x":<x_coordinate>,"y":<y_coordinate>}
{"action_type":"type","text":<text_input>,"x":<x_coordinate>,"y":<y_coordinate>}
{"action_type":"navigate_home"}
{"action_type":"navigate_back"}
{"action_type":"scroll","direction":<up, down, left, or right>}
{"action_type":"open_app","app_name":<app_name>}
{"action_type":"wait"}
{"action_type":"dismiss","x":<x_coordinate>,"y":<y_coordinate>}
{"action_type":"long_press","x":<x_coordinate>,"y":<y_coordinate>}
{"action_type":"get_text","x":<x_coordinate>,"y":<y_coordinate>}

If the goal of an instruction is reached, output the following special action
{"action_type":"status","goal_status":"successful"}
If the goal of an instruction is not possible, output the following special action
{"action_type":"status","goal_status":"infeasible"}

# Explanation of inputs
The top edge of a screen has y_coordinate equal to 0. The y_coordinate of the bottom edge of a screen equals to screen height. In screen_description, missing the 'checked' field for an element indicates that it is NOT checked.
The size of an element is defined by width and height.

# Input
instruction: {{{grounding_goal}}}

previous_actions: {{{previous_actions}}}
active_app: {{{active_app}}}
screen_width_height: {{{screen_width,screen_height}}}

screen_description: {{{screen_description}}}

The action to take:
\end{lstlisting}

\subsubsection{Few-shot prompt}
For a few-shot prompt, the following is inserted before "\# Input" in the \prompt prompt:
\begin{lstlisting}
Following are a few exemplars. Each exemplar is marked by <EXEMPLAR_i> and </EXEMPLAR_i> tags.
{{{exemplars}}}
\end{lstlisting}

The layout of an exemplar is as follows:
\begin{lstlisting}
<EXEMPLAR_{{{exemplar_index}}}>
# Input
instruction: {{{grounding_goal}}}

previous_actions: {{{previous_actions}}}
active_app: {{{active_app}}}
screen_width_height: {{{screen_width,screen_height}}}

screen_description: {{{screen_description}}}

The action to take:
{{{ground_truth_action}}}
</EXEMPLAR_{{{exemplar_index}}}>
\end{lstlisting}

\subsection{Action matching}
\label{relaxed_action_matching}
In computing step-wise accuracy, we consider an action correctly predicted if it matches the ground truth action exactly (i.e., action type and arguments are identical) or if it aligns with it as described below.

For element-based actions (click, long press, type), if the target element's coordinates are within the bounding box of the ground truth target element, it is considered as matching.
%a target coordinate is obtained either directly if the grounder predicts coordinates or computed as the center of the target element if the grounder predicts element indices. 
%This relaxation matches the behavior on Android devices where a touch gesture can penetrate all the elements containing the touch location and the proper element responses to the touch.
This relaxation matches the behavior on Android devices where a touch gesture will activate an element as long as it falls within the element's bounds.

%OR: I'm commenting this out because I'm worried it will raise lots of questions. If I understand what you're doing, you're applying some heuristics to extract the text and replace that into what the model predicted. You're basically bypassing the model and relying on heuristics which may not scale at all. We should avoid those heuristics and just rely on what the model predicts.
%For an action that requires a textual argument, such as \id{input\_text} and \id{open\_app}, if the ground-truth text argument can be parsed from the instruction, a predicted text is replaced by the closest ground-truth text. 

On Android the behaviour of the \id{navigate\_back} action is equivalent to clicking on the on-screen ``Back'' button so we consider them equivalent. Similarly, \id{open\_app} is considered equivalent to clicking a UI element whose text matches the app name.

\section{More experimental results}
\label{app:more-results}

\subsection{Confusion matrices for action predictions} 
\label{sec:confusion_matrix}

Table~\ref{tbl:confusion-matrix-hl} and \ref{tbl:confusion-matrix-ll} show the (normalized) confusion matrices with regard to action type predictions obtained with the LoRA-tuned PaLM 2S model (trained on the entire dataset). %A column corresponds to an action type that the grounder predicts while a row maps to a ground truth action type. Each row is normalized and sums to 1.0. Values on from rows cannot be compared directly as they are scaled differently. 
Results are averaged across all four \dataset's test splits. All the numbers are percentage. 

\begin{table}[h]
\centering
\caption{Confusion matrix for the action predictions of the LoRA-tuned PaLM 2S model (rank=64, trained on the entire training set) on high-level instructions.}
\vspace{-1ex}
\scalebox{0.85}{
\begin{tabular}{ll|cccccccc}
\multicolumn{2}{c}{} & \multicolumn{8}{c}{Predicted}\\ %\cline{2-9}
 & & click & input\_text & long\_press & navigate\_back & open\_app & scroll & wait & terminate\\
 \cline{2-10}
\multirow{8}{*}{\rotatebox{90}{Actual}}
& click & \textbf{88.1} & 0.6 & 0.1 & 1.5 & 0.0 & 4.6 & 3.8 & 1.3\\
& input\_text & 17.3 & \textbf{76.0} & 0.0 & 0.0 & 0.0 & 1.0 & 4.8 & 1.0\\
& long\_press & 28.6 & 0.0 & \textbf{71.4} & 0.0 & 0.0 & 0.0 & 0.0 & 0.0\\
& navigate\_back & 23.9 & 0.0 & 0.0 & \textbf{64.9} & 2.9 & 6.7 & 1.6 & 0.0\\
& open\_app & 2.9 & 0.0 & 0.0 & 2.6 & \textbf{93.3} & 0.6 & 0.6 &0.0\\
& scroll & 17.9 & 0.1 & 0.0 & 1.5 & 0.0 & \textbf{70.9} & 4.5 & 5.1\\
& wait & 14.1 & 0.3 & 0.0 & 1.6 & 0.0 & 10.8 & \textbf{68.6} & 4.6\\
& terminate & 16.0 & 0.6  & 0.0 & 1.5 & 0.3 & 30.7 & 12.2 & \textbf{38.6}\\
\cline{2-10}
\end{tabular}
\label{tbl:confusion-matrix-hl}
}
\end{table}

\begin{table}[h]
\centering
\caption{Confusion matrix of actions predictions of the LoRA-tuned PaLM 2S model (rank=64, trained on the all training set) on low-level instructions.}
\vspace{-1ex}
\scalebox{0.85}{
\begin{tabular}{ll|cccccccc}
\multicolumn{2}{c}{} & \multicolumn{8}{c}{Predicted}\\ %\cline{2-9}
 & & click & input\_text & long\_press & navigate\_back & open\_app & scroll & wait & terminate\\
 \cline{2-10}
\multirow{8}{*}{\rotatebox{90}{Actual}}
& click &  \textbf{93.3} & 0.3 & 0.0 & 0.2 & 0.0 & 0.8 & 3.7 & 1.7\\
& input\_text & 11.1 & \textbf{84.8} & 0.0 & 0.0 & 0.0 & 1.0 & 3.0 & 0.0\\
& long\_press & 12.5 & 12.5 & \textbf{75.0} & 0.0 & 0.0 & 0.0 & 0.0 & 0.0\\
& navigate\_back & 2.3 & 0.0 & 0.0 & \textbf{93.7} & 0.3 & 0.0 & 3.2 & 0.6\\
& open\_app & 0.2 & 0.0 & 0.0 & 0.0 & \textbf{98.5} & 0.0 & 1.3 &0.0\\
& scroll & 3.3 & 0.1 & 0.0 & 0.1 & 0.0 & \textbf{90.4} & 2.0 & 4.1\\
& wait & 11.8 & 0.2 & 0.2 & 1.4 & 0.0 & 4.6 & \textbf{76.0} & 5.8\\
& terminate & 10.5 & 1.3  & 0.0 & 0.8 & 0.6 & 7.0 & 12.4 & \textbf{67.5}\\
\cline{2-10}
\end{tabular}
\label{tbl:confusion-matrix-ll}
}
\end{table}

Actions of type \id{click} and \id{open\_app} are predicted with high accuracy (above 88\%). When UI actions are inferred from low-level instructions, performance is generally higher and mispredictions mainly occur for \id{long\_press} and \id{terminate} actions. Long press actions are not common in the dataset and in general in mobile apps, hence the model does not learn them as well as other actions. Task completion is generally hard to learn for the model and in many cases it is wrongly recognized as a pause. When UI actions must be inferred from high-level instructions, performance is naturally lower as the model must decompose the high-level instruction into a sequence of lower-level actions, thus requiring decision making and reasoning capabilities. The most challenging actions are \id{terminate}, \id{navigate\_back}, and \id{wait}. These actions are generally not explicit in a user instruction (e.g., a user may say ``download the file'' rather than ``download the file and wait for the download''), therefore requiring further reasoning and pre-knowledge of the task flow.

Please note that Table~\ref{tbl:confusion-matrix-hl} and \ref{tbl:confusion-matrix-ll} do not consider action arguments. Table~\ref{tbl:action_argument_accuracy} shows the accuracy of predicting all action arguments when the action type is predicted correctly.

\begin{table}[h]
\centering
\caption{Accuracy of predicting action arguments when the action type is predicted correctly. The numbers are in percentage. Results obtained with the LoRA-tuned PaLM 2S model (rank=64, trained on the all training set)} 
\vspace{-1ex}
\scalebox{0.9}{
\begin{tabular}{c|ccccccc}
 action type & click & input\_text & long\_press &  open\_app & scroll \\
\hline
high-level instructions & 76.2 & 75.9 & 60.0 & 85.6 & 90.6 \\
low-level instructions & 85.6 & 85.7 & 66.7 &  88.0 & 91.0  \\
\hline
\end{tabular}
\label{tbl:action_argument_accuracy}
}
\end{table}

The lowest accuracy is for \id{long\_press} actions which is most likely due to the scarcity of these actions in the dataset. Detecting the name of the target app works well as well as learning the direction of a scroll. In general, inferring action arguments, both target elements or input text, is much easier when the command is more explicit as in low-level instructions.

\subsection{Training with different levels of instructions}\label{sec:level_of_instructions}

As shown in Table \ref{tbl:grounding_accuracy_by_instruction_levels}, fine-tuning with a mixture of HL and LL SeqIO tasks of \dataset is equal or better than training individual models for different instruction levels. This is not a surprise as multi-task training increases the possibility of transfer learning. However, this may not always be true especially if different tasks contain conflicting data. 

\begin{table}[h]
    \caption{Step-wise accuracy (\%) on \dataset of a LoRA-tuned PaLM 2S model using various configurations of instructions for training and testing. Each column represents a model each trained on a SeqIO task or a mixture of SeqIO tasks, while each row corresponds to an evaluation SeqIO task. HL and LL stand for high-level and low-level instructions, respectively.}
    \centering
    \begin{tabular}{cccc}
    \toprule
       & trained on HL & trained on LL &  trained on HL+LL  \\
    \midrule
    tested on LL     & - & 83.4 &  \textbf{85.4}\\
    tested on HL     & \textbf{63.6} & -  &  \textbf{63.6}\\
    \bottomrule
    \end{tabular}
    \label{tbl:grounding_accuracy_by_instruction_levels}
\end{table}

\subsection{Random-500 vs. full test split}
\label{sec:random_500_vs_full_tests}

Table~\ref{tbl:random_500_vs_full_test} compares the difference of evaluating on Random-500 and the full test split. For zero-shot PaLM 2L, the step accuracy obtained on both splits are pretty similar, 35.2\% vs. 35.0\% for high-level instructions and 43.0\% vs. 42.7\% for low-level instructions. For fine-tuned PaLM 2S, the difference is larger but still smaller than the difference between a fine-tuned PaLM 2S and any zero-shot or few-shot model, so we consider it an accurate approximation for our analysis.

\begin{table}[h]
\centering
\caption{Grounding accuracy on Random-500 and the full test split. }
\vspace{-1ex}
\scalebox{0.90}{
\begin{tabular}{llccc}
\toprule
&  Model & Test split & \multicolumn{2}{c}{Step accuracy} \\
&           &  & high-level & low-level \\
 \midrule
Zero-shot & PaLM 2L & Random-500 & 35.2 & 43.0  \\
Zero-shot & PaLM 2L  & Full & 35.0 & 42.7 \\
\midrule
Fine-tuned & PaLM 2S & Random-500 & 62.6 & 82.2 \\
Fine-tuned & PaLM 2S  & Full &  64.8 & 80.0 \\

 \bottomrule
\end{tabular}
}
\label{tbl:random_500_vs_full_test}
\end{table}

\subsection{Step accuracy and episode accuracy vs. episode length}
\label{sec:accuracy_vs_episode_length}

% We perform further ablations whose results are in reported in Appendix~\ref{app:ablations}. In short, we observe that as tasks get longer, the probability of predicting all individual steps correctly decreases such that the best performing LoRA-tuned model we presented (PaLM-2S-LT-all-r64) can complete 5-step tasks only in 21\% of the cases and 6-step tasks only in 7.6\% of the cases. For the same task lengths its step accuracy is 71.3\% and 64.1\%, respectively (Figure~\ref{fig:hl-ep-len}). As previously noted, this demonstrates that a step accuracy well above 90-95\% is necessary for real-world scenarios. Certain action predictions (e.g., predicting a task is completed) are particularly challenging (see analysis in Appendix~\ref{sec:confusion_matrix}), but anyway necessary for the whole task to succeed.

\begin{figure}[t]
\centering
\begin{subfigure}{0.48\textwidth}
  \centering
    \includegraphics[width=\textwidth]{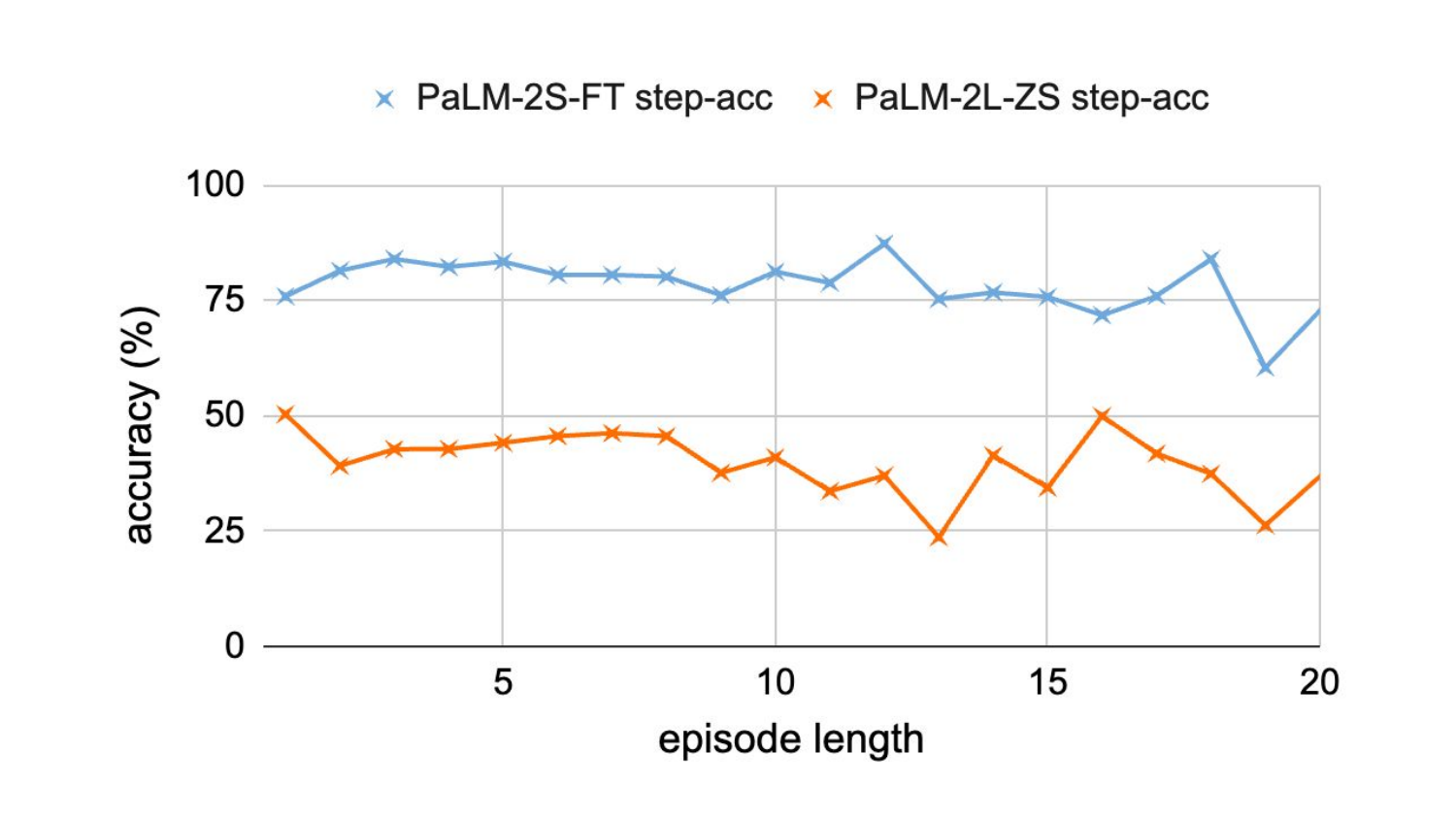}
    \caption{Low-level instructions}
    \label{fig:ll-ep-len}
\end{subfigure}
\hfill
\begin{subfigure}{0.47\textwidth}
  \centering
    \includegraphics[width=0.99\textwidth]{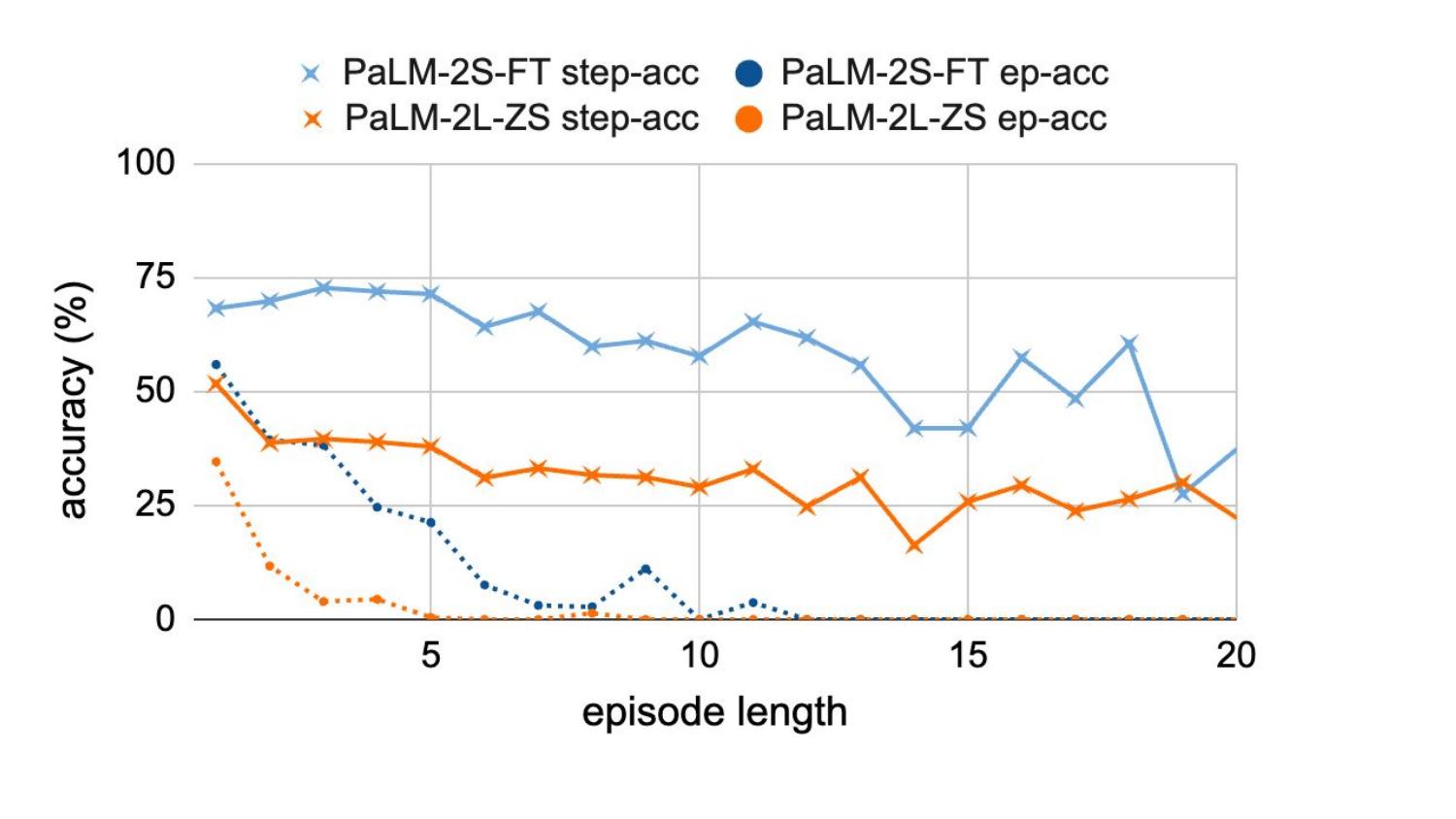}
    \caption{High-level instructions}
    \label{fig:hl-ep-len}
\end{subfigure}
\caption{Step accuracy and episode accuracy vs. episode length on the full test split. Tested models include PaLM-2S LoRA-tuned on the full dataset (PaLM-2S-FT) and PaLM-2L zero shot (PaLM-2L-ZS) using the \prompt prompt.}
\label{fig:accuracy_vs_episode_length}
\end{figure}

For this experiment we introduce the episode accuracy metric which measures the percentage of fully successful episodes. It is a harder metric since all step actions in a task must be predicted correctly for the task to be considered successful. We report this metric when testing on high-level instructions only, as for low-level instructions is less meaningful.
Figure~\ref{fig:accuracy_vs_episode_length} depicts how both step accuracy and episode accuracy vary when increasing the episode length from 1 to 20 steps. We report performance of the PaLM-2S model fine-tuned on the full dataset (PaLM-2S-FT, rank=64) and the PaLM-2L zero shot model using the \prompt prompt (PaLM-2L-ZS) on the full test split. The episode length has no impact on the step accuracy because the difficulty of a single step is independent on the episode length. As a result, this is relatively flat (see solid lines in both graphs in Figure~\ref{fig:accuracy_vs_episode_length}). However, as tasks become longer, the episode accuracy drops (dotted line in Figure~\ref{fig:hl-ep-len}). For the zero-shot model, tasks longer than 5 steps are never completed. For the fine-tuned model the drop is more gradual, but when going from 5-step-task to 6-step-task the episode accuracy drops from 21.3\% to 7.6\% (despite the corresponding step accuracy being 64--71\%). 

\subsection{Step-accuracy performance vs. application types}
\label{app:app-types}

%We observe the screen representations provided as input to the UI control agent are critical to its success. We break down the performance of some of the tested models based on whether the tested apps are Google apps or non-Google apps. 
%It is generally the case that Google proprietary apps (and high-trend apps) are more extensively annotated for accessibility, which leads UI control models to perform better on these apps, especially when given high-level instructions as input. As Table~\ref{tbl:first-party_vs_third-party} shows, on high-level instructions, the most best performing model (LT-all) achieves 82.5\% step accuracy on Google apps and only 58.7\% accuracy on non-Google apps.

We observe the screen representations provided as input to the UI control agent are critical to its success. Table~\ref{tbl:first-party_vs_third-party} compares the step accuracy of various models on Google's first party apps and apps developed by third party developers (3rd-party). It is evident the gap in performance, especially in handling high-level instructions. All zero-shot methods perform significantly better on first-party apps than on third-party apps. This is evident for fine-tuned models, especially on high-level instructions where the most performance model (LT-all) achieves 82.5\% step accuracy on first-party apps and only 58.7\% accuracy on third-party apps. When the input is made of low-level instructions the gap is smaller and in some cases (few-shots) the performance on third-party apps is higher. This shows how for tasks that require stronger reasoning capabilities accurate screen representations are particularly critical.

\begin{table*}[h]
\centering
\caption{Step accuracy performance of some of the zero-shot, few-shot, and LoRA-tuned models we studied  broken down by Google apps and non-Google (third party) apps. All tests are on Random-500. For all LoRA-tuned models rank=4 except for LT-all-r64.} 
\vspace{-1ex}
\scalebox{0.85}{
\begin{tabular}{lllcccc}
\toprule
Regime & Method & Model & \multicolumn{2}{c}{1st-party} &\multicolumn{2}{c}{3rd-party} \\
&  &           & hl & ll & hl & ll \\
\midrule
\multirow{4}{*}{Zero-shot} & \prompt & PaLM 2S  & 30.1 & 41.5 & 18.9 & 39.6 \\
 & \prompt & PaLM 2L & 39.8& 46.2 & 34.0 & 42.1 \\
 & \prompt & GPT-4  & \textbf{45.6} & \textbf{54.7} & 28.7 & 46.6 \\
& \prompt & Gemini 1.5  & 39.8 & 52.8 & 24.4 & 42.9 \\
%Zero-shot-AitW & PaLM 2L   & 34.0 & \textbf{54.7} & 15.1 & \textbf{51.0} \\
%Zero-shot-SeeAct & PaLM 2L  & 32.3  & 50.0   & 16.3 & 43.8 \\
\midrule
%Few-shot-2 & Gemini 1.0 & 42.7& 56.6 & 32.5 & 52.5\\
\multirow{3}{*}{Few-shot} & FS-5 & Gemini 1.5 Pro & \textbf{52.0} &37.7& 38.5& 52.0\\
& FS-10 & Gemini 1.5 Pro & 47.6 & 38.7 & 38.3 & 54.1\\
& FS-100 & Gemini 1.5 Pro & 45.6 & 49.1 & 37.9 & \textbf{54.5}\\
\midrule
%all use rank=4
\multirow{5}{*}{LoRA-tuned} & LT-5  & PaLM 2S   &  38.8 & 57.5 & 26.2 &56.3  \\ 
& LT-10 & PaLM 2S   & 36.9 &58.5&28.0&57.6 \\ 
& LT-100  & PaLM 2S  & 48.5 &59.4&35.8&61.9  \\ 
& LT-1k  & PaLM 2S  & 66.0&72.6&49.6&69.0  \\ 
& LT-10k & PaLM 2S & 74.8 & 79.2 & 54.7 & 77.9\\
& LT-all-64 & PaLM 2S & \textbf{82.5} & \textbf{86.8} & 58.7 & 85.0 \\ 
 \bottomrule
\end{tabular}
}
\label{tbl:first-party_vs_third-party}
\end{table*}

% \newpage
% \input{checklist}

\end{appendix}

\end{document}